%% file: bare_jrnl.tex
\documentclass[journal]{IEEEtran}
\pdfoutput =1
\usepackage{graphicx}
\usepackage{amsmath}
\usepackage{amssymb}
\usepackage{xfrac}
\usepackage{commath}
\usepackage{multirow}
\usepackage{enumitem}
\usepackage{booktabs}
\usepackage{xcolor}
\usepackage{array}
\usepackage{tabularx}

\usepackage{algorithm}
\usepackage{algpseudocode}
\usepackage{subcaption}

\ifCLASSINFOpdf
\else
\fi
\begin{document}
%
\title{Exploiting Context for Robustness to Label Noise in Active Learning}
%
%
%

\author{Sudipta Paul,~\IEEEmembership{Member,~IEEE}, Shivkumar Chandrasekaran, B.S. Manjunath,~\IEEEmembership{Fellow,~IEEE} and \\Amit K. Roy-Chowdhury,~\IEEEmembership{Fellow,~IEEE}
\thanks{$\bullet$ Sudipta Paul, and Amit~K.~Roy-Chowdhury are with the Department of Electrical and Computer Engineering, University of California, Riverside, CA, USA. Shivkumar Chandrasekaran and B.S. Manjunath are with the Department of Electrical and Computer Engineering, University of California, Santa Barbara, CA, USA.   \  E-mails: (spaul007@ucr.edu, shiv@ucsb.edu, manj@ucsb.edu,  amitrc@ece.ucr.edu)}}

%
%

\markboth{IEEE Transactions on Image Processing,~Vol.xx, No.xx, xxx~2020}%
{Sudipta \MakeLowercase{\textit{et al.}}: Bare Demo of IEEEtran.cls for IEEE Journals}
%



\maketitle

\begin{abstract}
\input{sections/N01_abstract}
\end{abstract}


\input{sections/N02_intro}
\input{sections/N03_related_work}
\input{sections/N04_methodology}

\input{sections/N05_experiment}
\input{sections/N06_conclusion}

%
\IEEEpeerreviewmaketitle

\section*{Acknowledgment}
The work was partially supported by ONR grant N00014-12-C-5113 and NSF grant 1901379

\ifCLASSOPTIONcaptionsoff
  \newpage
\fi



\bibliographystyle{IEEEtran}
\bibliography{Transactions-Bibliography/VCG_Group_pubs, Transactions-Bibliography/VCG_Group_Refs}

\end{document}

%% file: sections/N01_abstract.tex
Several works in computer vision have demonstrated the effectiveness of active learning for adapting the recognition model when new unlabeled data becomes available. Most of these works consider that labels obtained from the annotator are correct. However, in a practical scenario, as the quality of the labels depends on the annotator, some of the labels might be wrong, which results in degraded recognition performance. In this paper, we address the problems of i) how a system can identify which of the queried labels are wrong and ii) how a multi-class active learning system can be adapted to minimize the negative impact of label noise. Towards solving the problems, we propose a noisy label filtering based learning approach where the inter-relationship (context) that is quite common in natural data is utilized to detect the wrong labels. We construct a graphical representation of the unlabeled data to encode these relationships and obtain new beliefs on the graph when noisy labels are available. Comparing the new beliefs with the prior relational information, we generate a dissimilarity score to detect the incorrect labels and update the recognition model with correct labels which result in better recognition performance. This is demonstrated in three different applications: scene classification, activity classification, and document classification.

%% file: sections/N02_intro.tex
\section{Introduction}
Most of the current visual recognition tasks are performed by supervised learning approaches, which require a lot of training data. Every day a lot of visual and text data is generated from various sources which we can manually label and utilize to update the recognition system. But manually labeling a huge amount of data is tedious work and it becomes expensive if human experts are used. To reduce the labeling task, one effective approach is to actively select informative samples for manual labeling and update the recognition models with these selected samples. This scheme is known as active learning. As all the training samples may not be useful for developing a recognition system, active learning can reduce the labeling cost without compromising the recognition performance much \cite{li2014multi, S12, li2013adaptive, cuong2013active, kading2016large, sourati2018, zhang2020state, gudovskiy2020deep}.

In most of the active learning works, it is assumed that the labels provided by the human labelers are correct. However, in a practical scenario, labels queried from non-expert labelers are prone to error due to perception variations or incorrect annotation \cite{frenay2014classification}. Incorrect labels can adversely impact the classification performance of the influenced classifier \cite{Zhu2004noise}. This adverse impact becomes severe in an active learning process as the amount of labeled data is limited. There are some works \cite{hua2013collaborative, long2013active, long2016joint, long2015multi} that consider active learning where an annotator can provide wrong labels. Most of these works \cite{hua2013collaborative, long2013active, long2016joint} only consider label noise problem for binary classification. In contrast, visual recognition systems typically require to perform multi-class classification tasks. In \cite{long2015multi}, the authors studied multi-class multi-annotator active learning in the presence of label noise. They propose active learning with Robust Gaussian Process (RGP). However, as the computational cost of inference in Gaussian Process is $\mathcal{O}(n^3)$ \cite{wang2018partial}, this approach is not applicable to large scale datasets. Moreover, there can be many applications where it may not be possible to get multiple annotators, e.g., those that require a high level of domain knowledge. Hence, the problem of multi-class active learning in the presence of label noise requires more exploration. Furthermore, none of the previous approaches consider detecting which of the queried labels are wrong. Detection of wrong labels is of significant importance as it can be valuable for many applications like dataset creation with minimal human effort, annotator expertise estimation, and identifying samples that are difficult to annotate.  


\begin{figure*}
	\centering
	\includegraphics[scale=0.44]{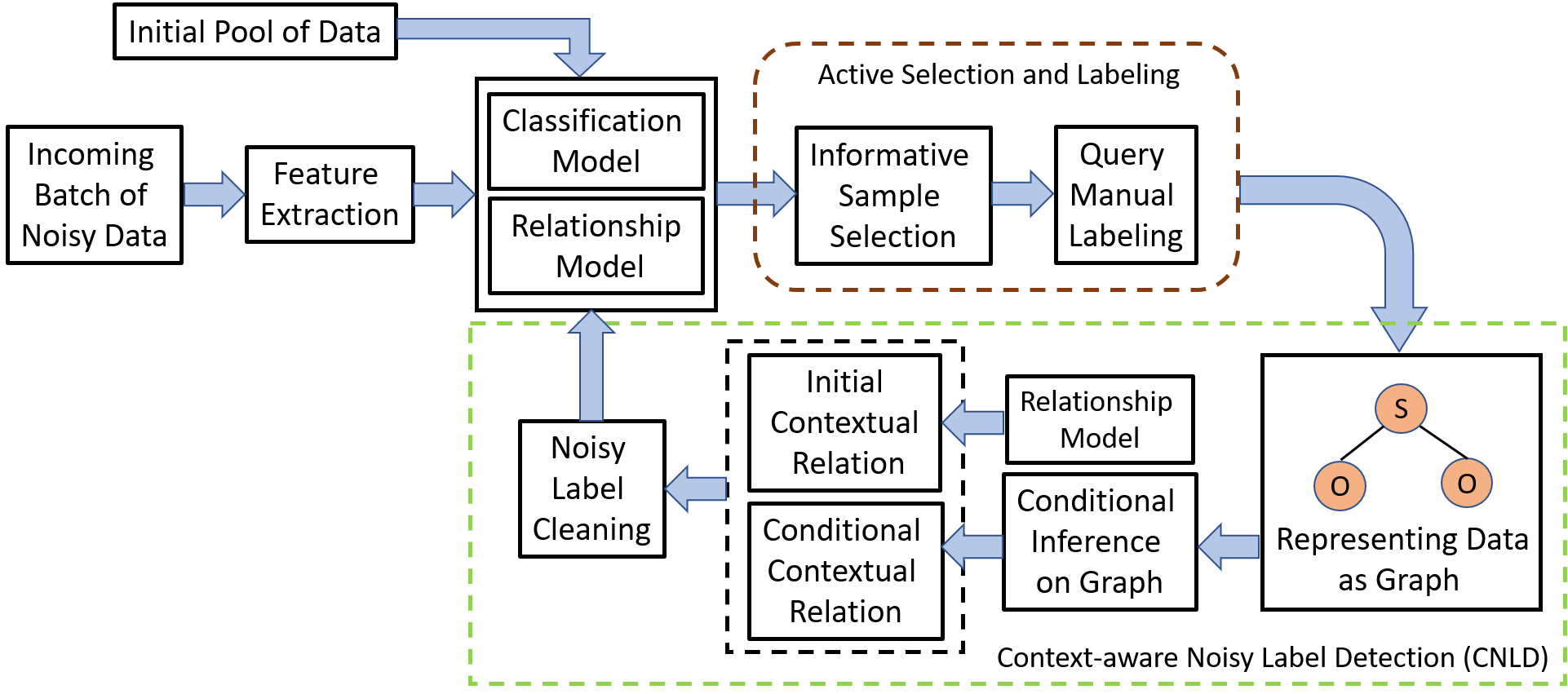}
	\caption{Proposed framework for label noise-robust active learning scheme. Initial classification model $\mathcal{M}^{t_0}$ and relationship model $\mathcal{R}^{t_0}$ are obtained using initial correctly labeled pool of data. When a new batch of data is available, it selects an informative subset of samples and queries for human labeling where a fraction of the queried labels is considered wrong. Graphical representations are constructed to encode contextual information. Conditional inference on the graph gives new edge beliefs which are compared with prior beliefs to obtain a measure of how likely a label is wrong. Then the classification model $\mathcal{M}^{t_0}$ and relationship model $\mathcal{R}^{t_0}$ are updated using only the correct labels.}
	\label{framework}
\end{figure*}

In this work, we propose a Context-aware Noisy Label Detection (CNLD) approach to detect wrong labels and utilize CNLD to formalize an active learning framework to handle the adverse impact of label noise. 
In many applications, several works have shown how to utilize the relationships between data points, i.e., structure in the data, for different purposes. 
For example, the relationship is used to improve recognition performance in activity recognition \cite{yao2010modeling}\cite{WSS13}, object recognition \cite{SUN397}\cite{galleguillos2008object}, and text classification \cite{bilgic2009link} \cite{settles2008analysis}. This inter-relationship is often termed as context. 
We also utilize the inter-relationships that are quite common in natural data to detect noisy labels. Generally, an incremental learning scenario that uses active selection has one or more initial seed models (classification model, relationship model) that are built on correct labels. We leverage the seed relationship model to obtain prior relational information among the data classes. When new data with queried labels are available, we infer its relational information using graphical representations, compare it with prior relations, and based on that obtain a dissimilarity score which is a notion of how likely an instance is incorrectly labeled. Utilizing this estimation, we detect which labels are wrong, filter out the wrong labels, and continue the learning process with correct labels. The motivation for this noisy label detection approach is that an instance assigned with the wrong label will lead to relational information among data classes that are not consistent with the known prior relational information among data classes.

\underline{\textit{Framework Overview.}}
Figure \ref{framework} illustrates the framework of our proposed approach. The framework is based on two assumptions: i) there is an underlying structure in the data which provides contextual relationships among the data classes and ii) we have an initial pool of data that is correctly labeled. Both of these are weak assumptions as almost all visual data occurring naturally is structured and most active learning methods starts with an initial seed model which is learned using a small set of labels queried from expert annotators. The method starts with training a classification model $\mathcal{M}$ and a relationship model $\mathcal{R}$ using the initial pool of correctly labeled data. When a new batch of unlabeled data is available, an informative subset of the samples is actively selected. These informative samples are queried for human labeling. We consider the practical scenario where a subset of the queried labels is wrong. Our goal is to improve the recognition performance by updating the classifier with queried labels.

Since the incorrect labels may confuse the learning process resulting in poor performance \cite{long2010rand}, we formulate a noisy label filtering based approach to reduce the influence of wrong labels during the learning process. To filter the noisy labels, we start by representing the queried instances along with their linked elements and defined attributes as graphs. Using the learned classifier, a probability mass function over the possible classes is obtained for each queried instance and its linked elements. We assign this probability mass function as node potentials. Edge potentials are obtained from the current relationship model. A message-passing algorithm is used to perform conditional inference. Conditional inference gives new edge beliefs which we consider as the posterior contextual information. We compare the posterior contextual information with prior contextual relation (obtained from the current relationship model) to compute a dissimilarity measure which is used to detect wrong labels. Then the classification model $\mathcal{M}$ and the relationship model $\mathcal{R}$ are updated using the filtered labels.

\underline{\textit{Main Contributions.}} The main contributions of the work are as follows.
\begin{itemize}[leftmargin=*]
    \item We derive a dissimilarity score to determine wrong labels by exploiting the inter-relationship among data categories, which is ubiquitous in natural data. 
    \item We formalize a general active learning framework that utilizes the inter-relationship based dissimilarity score to filter noisy labels provided by the annotator. 
    \item We empirically evaluate the performance of the Context-aware Noisy Label Detection (CNLD) approach, as well as its positive impact on active learning, on three different applications.
\end{itemize}

%% file: sections/N03_related_work.tex
\section{Related Works}

There has been a considerable amount of work on active learning. Most of the active learning algorithms use the uncertainty of the classifier as a measure of informativeness of an unlabeled data, e.g., entropy \cite{li2013adaptive}, best vs. second best \cite{li2012incorporating}, classifier margin \cite{vijayanarasimhan2014large}. Another common concept in active learning algorithms is expected model output change utilized in \cite{vezhnevets2012active,kading2016large,cai2013maximizing}. All of these methods consider the samples to be independent of each other. There have been some active learning methods that utilize relational information \cite{mac2014hierarchical, li2014multi, hasan2018context}. Recently some active learning methods are developed to scale well with deep learning network, e.g., core-set approach \cite{sener2017active}, deep Bayesian approach \cite{gal2017deep}, learning loss based approach \cite{yoo2019learning}, variational adversarial approach \cite{sinha2019variational}. 

There have been some works on active learning in the presence of label noise. Most of the works have a base setting where the learner is given an input space $\mathcal{X}$, a label space $\mathcal{L}$, and a hypothesis class $\mathcal{H}$. The goal is to select a hypothesis from the hypothesis class $\mathcal{H}$ which is closest to the hypothesis that generates the ground truth labels. Two types of noisy label settings are commonly used in these works: i) random classification Noise (RCN), where each label is flipped with a probability that is independent of the instances, and ii) increase of noise rate near the decision boundary. In \cite{kaariainen2006active}, the RCN setting is used and addressed by repeatedly querying an example. The sampling strategy in \cite{javidi2015noise} utilizes Extrinsic Jensen-Shanon (EJS) divergence. In \cite{castro2008minimax,dasgupta2008agnostic,beygelzimer2010agnostic, zhang2014beyond}, the second setting of noise where noise rate increases near decision boundary is studied. Works on agnostic active learning \cite{balcan2009agnostic, balcan2013active,beygelzimer2010agnostic, dasgupta2008agnostic, hanneke2007teaching, zhang2014beyond} considers a fraction of label may disagree with the optimal hypothesis of the hypothesis class $\mathcal{H}$. However, maintaining a hypothesis class may not be feasible for many computer vision applications.

There are some works on noise-robust active learning in the presence of multiple annotators. In \cite{hua2013collaborative}, active learning of kernel machine ensemble in collaborative labeling when labels might not be noise free is explored. In \cite{long2013active,long2016joint}, a noise resilient probabilistic model for active learning of a Gaussian process classifier from crowds is used. However, these approaches are designed for binary classification task. Long et al. \cite{long2015multi} have studied multi-class multi-annotator active learning using robust Gaussian process for visual recognition. In contrast, our work focuses on the presence of noisy labels for each annotator. 
Moreover, we are utilizing inter-relationship among data to detect the noisy labels 
and improve the robustness of the active learning framework. 

%% file: sections/N04_methodology.tex
\section{Methodology}

\underline{\textit{Problem Definition.}} Suppose, we have an initial set of data instances $\mathcal{L}$ that is correctly labeled. We extract features $\boldsymbol{X}^{\mathcal{L}}$ from this initial set of labeled data and train baseline classification model $\mathcal{M}$ and relationship model $\mathcal{R}$. Then a new batch of unlabeled data $\mathcal{U}$ consisting of $N$ data instances becomes available. We represent the extracted features of the unlabeled set of data as $\{\boldsymbol{X}_j^{\mathcal{U}}\}_{j=1}^N$. To update the classification model $\mathcal{M}$, an active sample selection procedure selects a subset $\mathcal{Q}$ of $k$ unlabelled data instances from the unlabeled set of data $\mathcal{U}$, where $k \le N$. This set of $k$ instances, $\mathcal{Q}=\{q_1,q_2,\dots,q_k\}$ is queried for manual labeling. After labeling by human annotator, we have the obtained labels $Y'=\{y'_1,y'_2,\dots,y'_k\}$. Considering the scenario where manual labeling is prone to error, we assume $\Omega$ fraction of the observed labels are wrong and the noise rate is unknown to the system. The true labels $Y=\{y_1,y_2,\dots,y_k\}$ of selected set $\mathcal{Q}$ is also unknown. Our goal here is to identify and remove the wrong labels and update the classification model $\mathcal{M}$ and relationship model $\mathcal{R}$ with only the correct labels so that the wrong labels cannot influence the models adversely.


\underline{\textit{Noisy Label Generation.}} We consider two statistical models to generate noisy labels synthetically. The Noisy Completely at Random (NCAR) \cite{frenay2014classification} statistical model is used to generate symmetric label noise and the Noisy at Random (NAR) \cite{frenay2014classification} model is used to generate asymmetric label noise. 

In the NCAR model, occurrence of an error has no relation with the instance or the label of that instance. Let the set of possible classes be $\mathcal{Z}=\{c_1,c_2,\dots,c_n\}$ for a set of data with $n$ classes and the selected noise rate is $\Omega$ for each class. Assigning equal noise rate $\Omega$ to each class results in symmetric noise. For the $i^{th}$ class, $\Omega$ fraction of randomly chosen labels are assigned with randomly chosen classes from the set $\mathcal{Z}\backslash\{c_i\}$ using the NCAR statistical model.

In the NAR model, occurrence of an error depends on the true label of the instance. We use this model to generate asymmetric noise as we can define which classes are more prone to label noise using the NAR model. Label noise generated by the NAR statistical model can be characterized by label transition matrix \cite{frenay2014classification}. Let two random variables $Y$ and $\Tilde{Y}$ denotes the true label and the assigned label respectively. So the label transition matrix characterizing the label noise noise generation is, \begin{equation}
\Lambda = \begin{pmatrix}
P(\Tilde{Y} =c_1|Y=c_1) &\hdots & P(\Tilde{Y} =c_n|Y=c_1)\\
\vdots &\ddots & \vdots\\
P(\Tilde{Y} =c_1|Y=c_n) &\hdots & P(\Tilde{Y} =c_n|Y=c_n)\\
\end{pmatrix}  
\label{eqn:transition}
\end{equation}

Here, $P(\Tilde{Y} = \tilde{y}|Y=y)$ is the label transition probabilities from true class to assign class. We use k-means clustering on the training data to obtain the transition probabilities, which is discussed in section (\ref{sec:noise_detection}).

\begin{figure}
	\centering
	\includegraphics[scale=0.38]{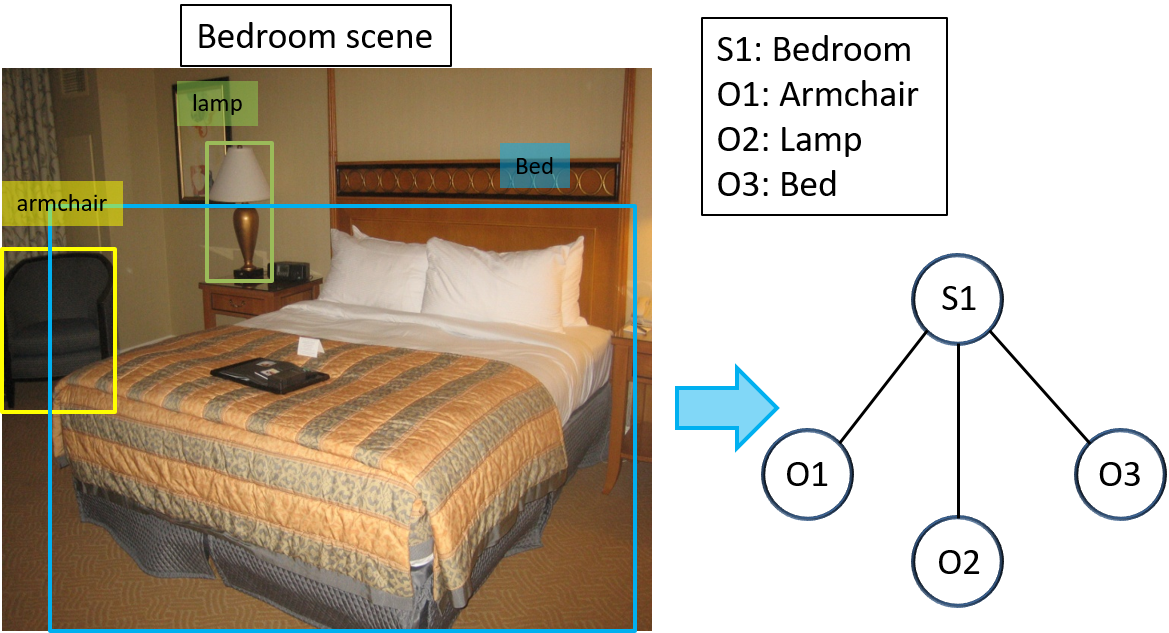}
	\caption{An example illustration of how the instances are represented as graphical structure. In the scene classification task, the shown image has a scene tag: bedroom and 3 objects: bed, armchair, and lamp. We represent the image by a tree structured graph with four nodes (one scene node and three object nodes) and three edges (scene-object).}
	\label{graph_representation_1}
\end{figure}

\subsection{Modeling Contextual Relationships}
\label{subsection:graph}
Inspired by the success of graphical representation in encoding contextual relationships in several applications \cite{yao2012describing, hasan2018context, SUN397}, we also utilize graphical representation to encode contextual relationships. Graph construction and how we define the node potentials and the edge potentials are described below.

\underline{\textit{Graph Formulation.}} We model the inter-relationship among the data by constructing an undirected graph $G=(V, E)$. Each node in $V$ represents a single instance. The edges $E=\{(i, j)|v_i$ and $v_j$ are linked$\}$ represent the relationships between the data points. If related attributes $\mathcal{A}$ are present in the structured data, we also model these into a graphical representation. Then the undirected graph $G=(V, E)$, modeling inter-relationship among the data points and also the relationships between the data points and related attributes $\mathcal{A}$ will have two types of node, $V=\{D, A\}$ and two types of edges, $E=\{D-D, D-A\}$. Here, $D$ represents the set of nodes corresponding to the data instances and $A$ represents the set of nodes corresponding to the related attributes. $D-D$ and $D-A$ are the relationships among data points and between data points and attributes respectively. Related attributes depend on the specific application, e.g., in scene classification, objects present in an image can be used as attributes. We consider the case of document classification where only the links between data points $(D-D)$ are present, scene classification where only the links between data points and related attributes $(D-A)$ are present and activity classification where both links between data points $(D-D)$ and links between data points and attributes $(D-A)$ are present. For example, in scene classification task, we represent data instance (scene) as scene node and detected objects as attribute nodes. As shown in Figure \ref{graph_representation_1}, we formulate a tree structured graph to represent the image.



\begin{algorithm}[ht]

    \caption{Context-Aware Noisy Label Detection (CNLD)}
	\label{algo:ir_framework} 
	\begin{algorithmic}
		\State {\bf Input:} 
		\begin{itemize}
		    \item Initial correct
		    classification model $\mathcal{M}^{t_0}$
		    \item Initial correct relationship model $\mathcal{R}^{t_0}$
		    \item Annotated noisy labels
		\end{itemize}
		\State {\bf Output:}
		Detection of wrong labels
		\State \textbf{step 1:} Calculate  $P(\mathcal{C}^D|c)$ and $P(\mathcal{C}^A|c)$  from $\mathcal{R}^{t_0}$
		\For{each element ($q_i$) of $\mathcal{Q}$}
		\State \textbf{step 2:} Construct $G_i = (V_i,E_i)$ 
		\State \textbf{step 3:} Calculate $\hat{P}(\mathcal{C}^D_i|c)$ and $\hat{P}(\mathcal{C}^A_i|c)$ by conditional inference
		\State \textbf{step 4:} Calculate dissimilarity score $l_i$ using Eqn. \ref{dis_score} 
		\EndFor
		\State \textbf{step 5:} Estimate weight $\gamma$ using Eqn. \ref{weight}
		\State \textbf{step 6:} Based on $\beta$ detect noisy labels.
    \end{algorithmic}
\end{algorithm}

\underline{\textit{Node Potential.}} Let us consider that we have a classification task where the data belongs to one of $n$ classes of $\{c_1, \dots, c_n \}$. Given a classifier $\mathcal{M}$, it can generate probability estimate of a data instance belonging to any class of $\{c_1, \dots, c_n \}$. The probability estimate of node $j$ belonging to some class $c_i$ can be expressed as $\mathcal{M}(\boldsymbol{X}_j,c_i)$. Consider an indicator function $\mathcal{I}(.)$ which takes as input a class $c$ and provides as output a unit standard basis vector, i.e., $\mathcal{I}(c=c_1)=[1, 0, \dots, 0]^T$. So the vector containing the node potentials of the $j^{th}$ node for $n$ classes can be expressed as, 
\begin{equation}
    \varphi_j = \sum_{i=1}^n
    \mathcal{I}(c=c_i)\mathcal{M}(\boldsymbol{X}_j,c_i)
\end{equation} 

\underline{\textit{Edge Potential.}} The edge potentials are obtained using the co-occurrence frequency \cite{galleguillos2008object}. Co-occurrence statistics can give an estimate of how likely two data classes are related or how data classes are related to the attribute classes. For example, in an image tagged with \lq bedroom' scene class, objects such as \lq bed' and \lq chair' are more likely to occur than a \lq car'. So the edge potentials represent the relationship weights among the data classes as well as between data classes and attribute classes. For two types of edges, we assign two different edge potential matrices $\boldsymbol{\Psi}_{D-D}$ and $\boldsymbol{\Psi}_{D-A}$. Here $\boldsymbol{\Psi}(i,j)$ is the co-occurrence frequency of class $c_i$ with class $c_j$. Calculation of co-occurrence frequencies are application-specific and discussed in section \ref{experiment}. 

\subsection{Context-aware Noisy Label Detection}\label{detection}
Suppose, a subset of data $\mathcal{Q}=\{q_1,q_2,\dots,q_k\}$, consisting of $k$ elements, is queried for human labeling. We consider that $\Omega$ fraction of the labels are incorrect. In this section, we discuss how we detect the incorrect labels using the graphical representations that encode the relationships among the data and also among data and attributes.

\underline{\textit{Contextual Relation.}} The relationship model $\mathcal{R}$ contains the co-occurrence information of different data classes and attribute classes. For a $n$ class classification task with classes $\{c_1,c_2,\dots,c_n\}$, we have $n \times n$ matrix $\boldsymbol{\Psi}_{D-D}$, where the $(i,j)^{th}$ value, $\boldsymbol{\Psi}_{D-D}(i,j)$ represents the co-occurrence statistics of data class $c_i$ and data class $c_j$. Using this co-occurrence information we can calculate the probability of presence of $i^{th}$ data class $c_i$ in presence of $j^{th}$ data class $c_j$ by,

\begin{equation}
    \label{p_data}
    P(c_i|c_j)=\frac{\boldsymbol{\Psi} _{D-D}(j,i)}{\sum_{i=1}^n \boldsymbol{\Psi}_{D-D}(j,i)}
\end{equation}

Similarly, from the relationship model $\mathcal{R}^{t_0}$, we have the co-occurrence frequency of data classes and attribute classes $\boldsymbol{\Psi}_{D-A}$. If there are $m$ attribute classes $\{a_1,a_2,\dots,a_m\}$, the probability of presence of $i^{th}$ attribute class $a_i$ in presence of $j^{th}$ data class $c_j$ can be expressed by,

\begin{equation}
    \label{p_attri}
    P(a_i|c_j)=\frac{\boldsymbol{\Psi}_{D-A}(j,i)}{\sum_{i=1}^n \boldsymbol{\Psi}_{D-A}(j,i)}   
\end{equation}




\underline{\textit{Prior Relational Information.}} From the current relationship model $\mathcal{R}^{t_0}$, we know $\boldsymbol{\Psi}_{D-D}$ and $\boldsymbol{\Psi}_{D-A}$, which we consider as the prior edge beliefs. Suppose, in an $n$ class classification problem with $m$ classes of attributes, $\mathcal{C}^D$ is a random variable with sample space $\{c_1,c_2,\dots,c_n\}$ and $\mathcal{C}^A$ is a random variable with sample space $\{a_1,a_2,\dots,a_m\}$. Using the prior edge beliefs in Eqn \ref{p_data} and \ref{p_attri}, we obtain the conditional distribution $P(\mathcal{C}^D|c)$ and $P(\mathcal{C}^A|c)$, which we call the prior relational information for instances of $\mathcal{Q}$. 

\underline{\textit{Dissimilarity Score Generation.}} 
In order to detect the wrong labels by exploiting the relationships, we construct graphical representation as described in \ref{subsection:graph} for each element of $\mathcal{Q}$ along with their linked data instances and attributes. Consider an instance $q$ is linked with $e$ data instances and attributes and the queried label is $y'$. We construct graph for $q$, which can be represented as $G=(V,E)$ where $V=\{v_1,v_2,\dots,v_{e+1}\}$ and $E=\{(1,j)|j=2,\dots,e+1\}$. Here the node $v_1$ represents the data instance $q$ and nodes $v_2,...,v_{m+1}$ are linked data instances and attributes of $q$. The graph has $e$ edges that connects $v_1$ with all the other nodes. So $G$ forms a tree structure. The node potentials $\varphi$ and the edge potentials $\boldsymbol{\Psi}_{D-D}$, $\boldsymbol{\Psi}_{D-A}$ are assigned using the classification model $\mathcal{M}^{t_0}$ and relationship model $\mathcal{R}^{t_0}$.

Now for each instance of $\mathcal{Q}$, we estimate its class conditional relatedness with other classes by making conditional inference on the representative graph. Conditional inference gives the pairwise conditional distribution of classes for each edge, which we call the posterior edge beliefs $\hat{\boldsymbol{\Psi}}$. Using the posterior edge beliefs of all edges in a graph, we estimate the posterior probability distribution conditioned on a class for each instance of $\mathcal{Q}$.


Suppose, we assign $j^{th}$ class $c_j$ to the instance $q$ and make conditional inference on graph $G$. Let the number of $D-D$ edges is $e^1$ and the number of $D-A$ edges is $e^2$ in graph $G$. We estimate the posterior probabilities by, 
\begin{equation} 
    \hat{P}(c_i|c_j)= \frac{1}{e^1} \frac{\sum_{k=1}^{e^1} \hat{\boldsymbol{\Psi}}_{(D-D)_k}(j,i)}{\sum_{k=1}^{e^1}\sum_{i=1}^{n} \hat{\boldsymbol{\Psi}}_{(D-D)_k}(j,i)}, 
    \label{p_data_pos}
\end{equation}

\begin{equation} \label{p_attri_pos}
    \hat{P}(a_i|c_j)=\frac{1}{e^2} \frac{\sum_{k=1}^{e^2} \hat{\boldsymbol{\Psi}}_{(D-A)_k}(j,i)}{\sum_{k=1}^{e^2}\sum_{i=1}^{m} \hat{\boldsymbol{\Psi}}_{(D-A)_k}(j,i)}, 
\end{equation}

where $\hat{\boldsymbol{\Psi}}_{(D-D)_k}$ represents posterior edge beliefs of $k^{th}$ $D-D$ edge and $\hat{\boldsymbol{\Psi}}_{(D-A)_k}$ represents posterior edge beliefs of $k^{th}$ $D-A$ edge. Using Eqn \ref{p_data_pos} and \ref{p_attri_pos}, we obtain the posterior conditional distribution $\hat{P}(\mathcal{C}^D|c)$ and $\hat{P}(\mathcal{C}^A|c)$ for each instance of $\mathcal{Q}$, which we call the posterior relational information.


We rely on the idea that an instance is most likely wrongly labeled by the annotator if the posterior relational information for the assigned class is not consistent with the prior relational information for that class. We use Kullback-Leibler divergence on the prior and posterior relational information and use that to assign a dissimilarity  score $L=\{l_1,l_2,\dots,l_k\}$ on each element of $\mathcal{Q}$. This dissimilarity score gives a measure of how dissimilar the prior and posterior relational information is. If $i^{th}$ element $q_i$ is labeled with $k^{th}$ class $c_k$ by the annotator, we assign the dissimilarity score to the $i^{th}$ instance by,

\begin{align}
    \footnotesize
    \label{dis_score}
    l_i= \frac{1}{n} \sum_{j=1}^n \max & \Big( D_{KL} (\hat{P}(\mathcal{C}^D_i|c_k)||P(\mathcal{C}^D|c_k)) \nonumber
    \\ 
    & - D_{KL} (\hat{P}(\mathcal{C}^D_i|c_j)||P(\mathcal{C}^D|c_j)),0 \Big) \nonumber
    \\
    & \hspace{-1.5cm} + \frac{1}{m} \sum_{j=1}^m \max \Big(D_{KL} (\hat{P}(\mathcal{C}^A_i|c_k)||P(\mathcal{C}^A|c_k)) \nonumber
    \\ 
    & - D_{KL} (\hat{P}(\mathcal{C}^A_i|c_j)||P(\mathcal{C}^A|c_j)),0 \Big)
 \end{align}

Here, $\hat{P}(\mathcal{C}_i^D|c)$ and $\hat{P}(\mathcal{C}_i^A|c)$ represent the posterior relational information of the $i^{th}$ instance $q_i$.  
 

\subsection{Model Update}
\label{sec:MU}
For $i^{th}$ labeled instance, the estimated weight from the dissimilarity score is,
\begin{equation}
    \gamma_i=1-\frac{l_i}{max(l_1,l_2,\dots,l_k)}
    \label{weight}
\end{equation}

We consider a threshold $\beta$ for detecting wrong labels. Instances for which $\gamma>\beta$, are considered correctly labeled. The classification model $\mathcal{M}^{t_0}$ and relationship model $\mathcal{R}^{t_0}$ is updated with these labels, which result in new classification model $\mathcal{M}^{t_1}$ and new relationship model $\mathcal{R}^{t_1}$.

%% file: sections/N05_experiment.tex
\section{Experiments} \label{experiment}
To evaluate the effectiveness of our proposed method, we conduct experimental analysis considering noisy labels in three different application domains: scene classification, activity classification, and document classification. These domains are selected as the data representative of the domains can share relationships among them, which is required to form the relationship model. 

\begin{table*}
\centering
\caption{Comparison of the performance of CNLD with other approaches for the noisy label detection task in symmetric label noise scenario. We compare the performance for the set of noise ratio $\Omega \in \{0.10,0.20,0.30,0.40,0.50\}$. The performance is evaluated in terms of Type-I error (ER1), Type-II error (ER2), and Noise Elimination Precision (NEP) for three datasets.}
\begin{tabular}{c|c|p{.45cm}p{.45cm}p{.45cm}|p{.45cm}p{.45cm}p{.45cm}|p{.45cm}p{.45cm}p{.45cm}|p{.45cm}p{.45cm}p{.45cm}|p{.45cm}p{.45cm}p{.45cm}}
\hline
\multirow{2}{*}{Dataset} & \multirow{2}{*}{Method} & \multicolumn{3}{c|}{\underline{$\Omega = 0.10$}} & \multicolumn{3}{c|}{\underline{$\Omega = 0.20$}} & \multicolumn{3}{c|}{\underline{$\Omega = 0.30$}} & \multicolumn{3}{c|}{\underline{$\Omega = 0.40$}} & \multicolumn{3}{c}{\underline{$\Omega = 0.50$}}\\
& & ER1 & ER2 & NEP & ER1 & ER2 & NEP & ER1 & ER2 & NEP & ER1 & ER2 & NEP & ER1 & ER2 & NEP \\
\hline
\hline
\multirow{4}{*}{Scene} & Majority \cite{brodley1999identifying} & 0.10 & 0.88 & 0.12 &
0.19 & 0.76 & 0.24 & 0.29 & 0.68 & 0.32 & 0.37 & 0.57 &
0.43 & 0.47 & 0.45 & 0.55\\
& Consensus \cite{brodley1999identifying} & 0.10 & 0.87 & 0.13 & 0.19 & 0.76 & 0.24 & 0.27 & 0.64 & 0.36 & 0.35 & 0.53 & 0.47 & 0.43 & 0.41 & 0.59\\
& Probabilistic \cite{sun2007identifying} & 0.09 & 0.81 & 0.19 & 0.19 & 0.76 & 0.24 & 0.26 & 0.61 & 0.39 & 0.34 & 0.52 & 0.48 & 0.43 & 0.43 & 0.58\\
& \textbf{CNLD} & \textbf{0.08} & \textbf{0.72} & \textbf{0.28} & \textbf{0.12} & \textbf{0.49} & \textbf{0.51} & \textbf{0.19} & \textbf{0.44} & \textbf{0.55} & \textbf{0.25} & \textbf{0.38} & \textbf{0.62} & \textbf{0.30} & \textbf{0.31} & \textbf{0.71} \\
\hline
\hline
\multirow{4}{*}{Activity} & Majority \cite{brodley1999identifying} & 0.09 & 0.80 & 0.20 & 0.16 & 0.63 & 0.37 & 0.22 & 0.51 & 0.49 & 0.26 & 0.40 & 0.60 & 0.31 & 0.31 & 0.69 \\
& Consensus \cite{brodley1999identifying} & 0.08 & 0.71 & 0.19 & 0.14 & 0.54 & 0.46 & 0.18 & 0.41 & 0.59 & 0.20 & 0.31 & 0.69 & 0.24 & 0.23 & 0.77 \\
& Probabilistic \cite{sun2007identifying} & 0.05 & 0.49 & 0.51 & 0.10 & 0.38 & 0.62 & 0.12 & 0.29 & 0.71 & 0.17 & 0.25 & 0.75 & 0.19 & 0.19 & 0.81 \\
& \textbf{CNLD} & \textbf{0.05} & \textbf{0.45} & \textbf{0.56} & \textbf{0.09} & \textbf{0.35} & \textbf{0.65} & \textbf{0.12} & \textbf{0.29} & \textbf{0.71} & \textbf{0.16} & \textbf{0.24} & \textbf{0.76} & \textbf{0.17} & \textbf{0.18} & \textbf{0.83}\\
\hline
\hline
\multirow{4}{*}{Document} & Majority \cite{brodley1999identifying} & 0.09 & 0.79 & 0.21 & 0.16 & 0.64 & 0.36 & 0.22 &
0.50 & 0.50 & 0.27 &
0.40 & 0.60 & 0.30 &
0.30 & 0.70\\
& Consensus \cite{brodley1999identifying} & 0.08 & 0.79 & 0.21 & 0.14 & 0.57 & 0.43 & 0.18 & 0.42 & 0.58 & 0.22 & 0.40 & 0.60 & 0.25 & 0.24 & 0.76 \\
& Probabilistic \cite{sun2007identifying} & 0.06 & 0.54 & 0.46 & 0.11 & 0.41 & 0.58 & 0.15 & 0.36 & 0.65 & 0.19 & 0.28 & 0.72 & 0.22 & 0.23 & 0.78 \\
& \textbf{CNLD} & \textbf{0.05} & \textbf{0.45} & \textbf{0.55} & \textbf{0.09} & \textbf{0.34} & \textbf{0.65} & \textbf{0.11} & \textbf{0.28} & \textbf{0.73} & \textbf{0.15} & \textbf{0.22} & \textbf{0.78} & \textbf{0.18} & \textbf{0.19} & \textbf{0.82} \\
\hline
\end{tabular}
\label{table:noise_detection}
\end{table*}

\subsection{Dataset} MIT-67 Indoor \cite{MITIndoor2009} dataset is used for scene classification application domain. The dataset contains images of $67$ indoor scene categories. We use the proposed split of trainset and testset by \cite{MITIndoor2009} where the trainset contains $80$ images per class and the testset contains $20$ images per class. For activity classification, we use VIRAT \cite{oh2011large} dataset. It consists of $329$ sequences of $11$ activity classes totaling $1422$ activities. For document classification, we use the CORA \cite{sen2008collective} dataset. It consists of $2708$ scientific publications divided into $7$ classes. These publications are linked by citations.

\subsection{Features and Graphical Representation}
\label{FGR}

\underline{\textit{Scene Classification.}} Scene features of MIT-67 Indoor dataset images are extracted using ResNet-50 \cite{he2016deep} pre-trained on the Places365 \cite{zhou2018places} dataset. An off-the-shelf object detector is used to detect objects that are present in the image. We use the Matterport Mask R-CNN implementation \cite{matterport_maskrcnn_2017_2}, which is built on Feature Pyramid Network (FPN) \cite{lin2017feature} and a ResNet101 \cite{he2016deep} backbone. We use a model trained on MS COCO dataset \cite{lin2014microsoft} to detect objects. Each image in the dataset is graphically represented by a single scene node and multiple object nodes corresponding to the objects detected in that image. Scene node potentials are obtained using the current scene classification model, which we learn incrementally with each incoming batch. Object node potentials are obtained using an off-the-shelf detector. To detect mislabeled scene nodes, we use the scene-object (S-O) relationships. All the object nodes are connected to the scene node in the graphical representation of an image forming a tree structure. We use the co-occurrence frequencies of scene classes and object classes to build the relationship model and assign edge potentials of the graph.

\underline{\textit{Activity Classification.}} C3D \cite{tran2014learning} model trained on sports-1M \cite{KarpathyCVPR14} dataset is used to extract features from activity segments. We extract the C3D features for every $16$ frames, with a temporal stride of eight frames, and apply max-pooling to obtain a feature vector $\boldsymbol{X}_j \in \mathbb{R}^{4096}$ for each segment. Each sequence of the VIRAT dataset is represented by an undirected graph. We consider two sets of nodes: activity and object/person and two sets of edges: activity-activity and activity-object/person. We consider activities within a certain spatio-temporal distance to be related to each other. The co-occurrence frequencies of the activity-activity or activity-object/person within a certain spatio-temporal region are used to build the relationship model and assign the edge potentials. Activity node potentials are obtained using the current activity classifier. Object/person node potentials are obtained using the same binning approach used in \cite{hasan2015context}.

\underline{\textit{Document Classification.}} In the CORA dataset, every publication instance is represented using a dictionary of $1433$ unique words. The feature vector $\boldsymbol{X}_j \in \{0,1\}^{1433}$ indicates the absence or presence of these words. We use the citation link information to build the graphs. Every publication instance is considered as a node and edges are formed when an instance is linked with another one via a citation. Node potentials are obtained using the current document classifier. Edge potentials are represented by a matrix consisting of the number of times a document of a certain class is linked to a document of another class. We also use this link information to build the relationship model.

\subsection{Experimental Setups}
\label{exp_setup} We conduct experiments to analyze both the performance of our proposed Context-aware Noisy Label Detection approach for detecting wrong labels and the proposed active learning framework for robust classification in all three application domains. We use two different experimental setups to analyze these two different tasks. Multinomial Logistic Regression (MLR) is used as the classifier for all three applications. Note that the steps of our algorithm are independent of the particular choice of a classifier. Publicly available UGM Toolbox \cite{schmidtugm} is used to infer on the node and edge beliefs. For all classification tasks, we divide the training set into multiple batches and these batches are made available sequentially. All the performance evaluations reported here are an average of multiple rounds of experiments. For scene classification, we use the proposed split by \cite{MITIndoor2009} and in each round of an experiment, we form batches of unlabelled data by randomly shuffling instances. We also change the set of instances that are assigned with wrong labels randomly in each round of an experiment. For activity classification, 176 sequences (761 activities) are used for training and other 153 sequences (661 activities) are used for testing. In each round of an experiment, we assign wrong labels to a randomly selected set of instances. For document classification, we use 10-fold cross-validation and different sets of instances are selected and assigned wrong labels in each round of an experiment.

\subsection{Noisy Label Detection Performance Analysis} \label{sec:noise_detection} In our proposed approach, we utilize context information to detect noisy labels. The context information is encoded using a graphical representation. For noisy label detection, we divide the training set in multiple batches and use the instances from a single batch to train an initial classification model $\mathcal{M}$ and an initial relationship model $\mathcal{R}$. We consider the labels of these instances are correct. In the testset, a fraction of the instances is assigned with wrong labels, where the wrong labels are generated synthetically. To detect the wrong labels in the testset, we represent the instances from the testset graphically as described in section \ref{FGR}. We utilize the trained models and the graphical representations to calculate the dissimilarity scores, where the dissimilarity score is an indicator of how likely a label is wrong. The detection of a noisy label can be considered as a binary classification problem and the normalized dissimilarity score from Eqn. \ref{dis_score} can be treated as the confidence score for this binary classification task. To analyze the performance of this proposed approach, we conduct the following experiments:
\begin{itemize}[leftmargin=*]
    \item We analyze the performance of CNLD for detecting wrong labels when symmetric label noise is considered.
    \item We analyze the performance of CNLD for detecting wrong labels when asymmetric label noise is considered.
\end{itemize}

To evaluate the performance of noisy label detection, we use Type-I error (ER1), Type-II error (ER2), and Noise Elimination Precision (NEP) as described in \cite{frenay2014classification}. Type-I error represents the percentage of correctly labeled instances that are erroneously removed. Type-II error represents the percentage of mislabeled instances which are not removed. Noise Elimination Precision measures the percentage of removed samples that are actually mislabelled. The corresponding measures are:
\begin{flalign*}
    \text{ER1 = } \frac{\text{\# of correctly labelled instances which are removed}}{\text{\# of correctly labelled instances}} 
\end{flalign*}
\begin{flalign*}
    \text{ER2 = } \frac{\text{\# of mislabelled instances which are not removed}}{\text{\# of mislabelled instances}} 
\end{flalign*}
\begin{flalign*}
    \text{NEP = } \frac{\text{\# of mislabeled instances which are removed}}{\text{\# of removed instances}} 
\end{flalign*}

\underline{\textit{Performance of CNLD for Symmetric Label Noise.}} Table \ref{table:noise_detection} illustrates the performance of CNLD for noisy label detection task in symmetric noise scenario. In this experimental setup, the noisy labels are generated synthetically using the Noisy Completely at Random (NCAR) statistical model. We consider symmetric noise where $\Omega$ fraction of each class of the testing set samples are assigned with wrong labels and consider $\Omega \in \{0.10,0.20,0.30,0.40,0.50\}$. We compare our proposed approach with majority voting \cite{brodley1999identifying}, consensus voting \cite{brodley1999identifying}, and probabilistic approach \cite{sun2007identifying}. Majority voting and consensus voting approaches use multiple classifiers to detect noisy labels. A label is detected as wrong if predictions of the majority of the classifiers disagree with the assigned label in the majority voting approach. Similarly, a label is detected as wrong if predictions of all of the classifiers disagree with the assigned label in the consensus voting approach. We use logistic regression, SVM, and kNN as classifiers for both of these approaches. All the classification models are learned using the same initial batch of correctly labeled data. In the probabilistic approach, the wrong labels are detected using the mismatch of assigned labels with classifier prediction and entropy of the class prediction. 

In our experimental setup, we detect $\Omega$ fraction of the testing set as wrong labels for all approaches and compute Type-I error (ER1), Type-II error (ER2), and Noise Elimination Precision (NEP) scores. Here, low ER1 and ER2 scores and high NEP scores indicate better performance. For all three application domains and different noise rate $\Omega$, there is a significant improvement in performance for our proposed approach. We observe a maximum of $26\%$, $5\%$, and $9\%$ absolute improvement in NEP scores in scene, activity, and document dataset respectively. In the scene dataset, NEP scores for majority voting and consensus voting approach are close to the value of noise rate ($\Omega$), indicating the inefficacy of the two approaches for noisy label detection. Compared to the activity and document dataset, we observe that the difference between noisy label detection performance of the baseline approaches and CNLD in the scene dataset is more significant. This is because, the three compared noisy label detection approaches solely rely on the performance of the learned classifiers, while CNLD relies on both the classifier and the context information. In the scene dataset, the learned classifier has less accuracy compared to the activity and document dataset, resulting in poor detection performance for the three compared approaches while CNLD retains a good detection performance by utilizing contextual information.

\underline{\textit{Performance of CNLD for Asymmetric Label Noise.}} Table \ref{table:noise_detection2} illustrates the performance of CNLD for noisy label detection task in an asymmetric noise scenario. We use the Noisy at Random (NAR) model to generate asymmetric label noise synthetically. The transition probabilities of the label transition matrix $\Lambda$ are calculated using k-means clustering. For a dataset with $n$ classes, we initialize $n$ cluster centers, where each cluster center is the calculated average of features from all the samples of a class. Then we use the assignment step and update step to update the cluster centers until convergence. We consider a cluster to be representative of a particular class if it contains most samples from that class. If a cluster represents class $y$, then we calculate the transition probabilities $P(\Tilde{Y} = \tilde{y}|Y=y)$ based on the number of samples that cluster contains from class $\Tilde{y}$. We obtain $10\%$, $7\%$, and $12\%$ absolute improvement in NEP scores compared to the best performing baseline approach in scene, activity, and document dataset respectively. We observe that for asymmetric label noise, the difference in performance between baseline approaches and our proposed approach is large compared to the symmetric noise scenario in activity classification and document classification. This indicates that compared to other approaches, CNLD is able to retain the detection performance in the asymmetric case.

\begin{table}
\centering
\caption{Comparison of the performance of CNLD with other approaches for the noisy label detection task in asymmetric label noise scenario. We observe an improvement of performance for our proposed CNLD in all three datasets.}
\begin{tabular}{c|c|ccc}
\hline
Dataset & Method & ER1 & ER2 & NEP\\
\hline
\hline
\multirow{4}{*}{Scene} & Majority \cite{brodley1999identifying} & 0.20 & 0.77 & 0.23\\
& Consensus \cite{brodley1999identifying} & 0.20 & 0.75 & 0.25\\
& Probabilistic \cite{sun2007identifying} & 0.19 & 0.73 & 0.27\\
& \textbf{CNLD} & \textbf{0.17} & \textbf{0.63} & \textbf{0.37}\\
\hline
\hline
\multirow{4}{*}{Activity} & Majority \cite{brodley1999identifying} & 0.27 & 0.41 & 0.59\\
& Consensus \cite{brodley1999identifying} & 0.21 & 0.32 & 0.68\\
& Probabilistic \cite{sun2007identifying} & 0.26 & 0.31 & 0.69\\
& \textbf{CNLD} & \textbf{0.16} & \textbf{0.24} & \textbf{0.76}\\
\hline
\hline
\multirow{4}{*}{Document} & Majority \cite{brodley1999identifying} & 0.24 & 0.41 & 0.59\\
& Consensus \cite{brodley1999identifying} & 0.21 & 0.36 & 0.64\\
& Probabilistic \cite{sun2007identifying} & 0.18 & 0.30 & 0.70\\
& \textbf{CNLD} & \textbf{0.10} & \textbf{0.18} & \textbf{0.82}\\
\hline
\end{tabular}
\label{table:noise_detection2}
\end{table}

\begin{figure*}
	\centering
	\begin{subfigure}{0.3\textwidth}
		\includegraphics[scale=0.38]{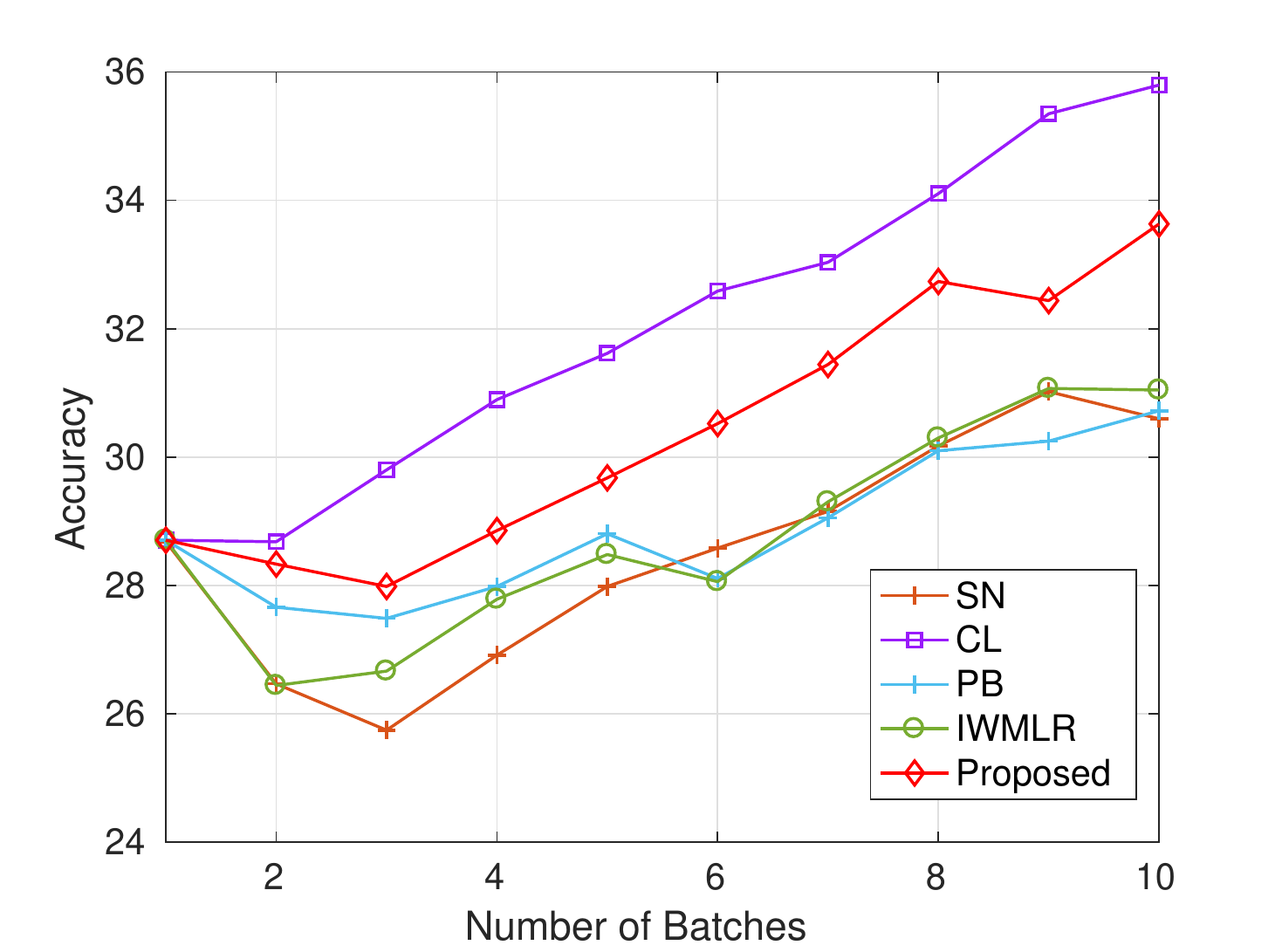}
		\caption{Scene}
		\label{acc_40_1}
	\end{subfigure}
	\begin{subfigure}{0.3\textwidth}
		\includegraphics[scale=0.38]{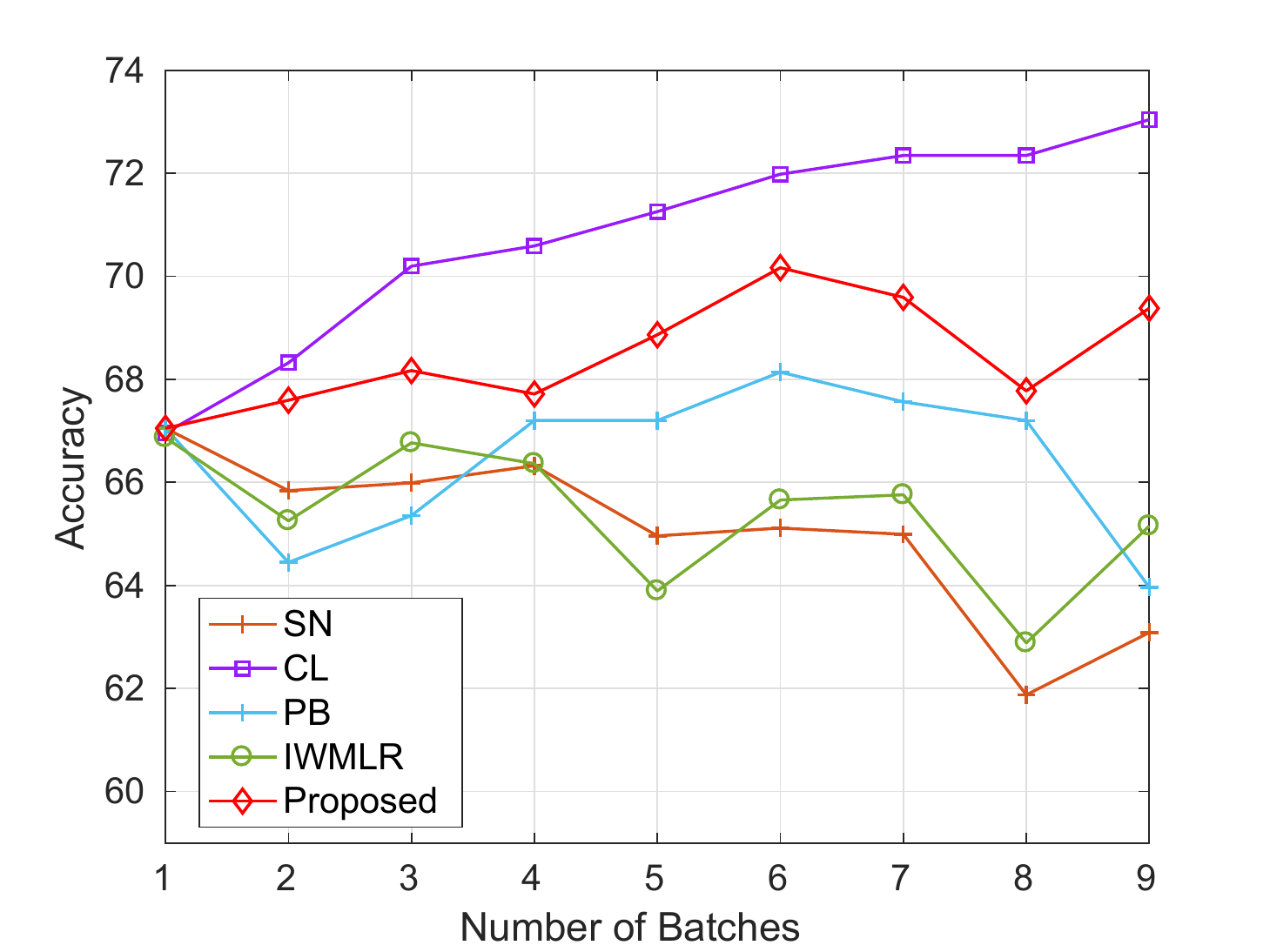}
		\caption{Activity}
		\label{acc_40_2}
	\end{subfigure}
	\begin{subfigure}{0.3\textwidth}
		\includegraphics[scale=0.38]{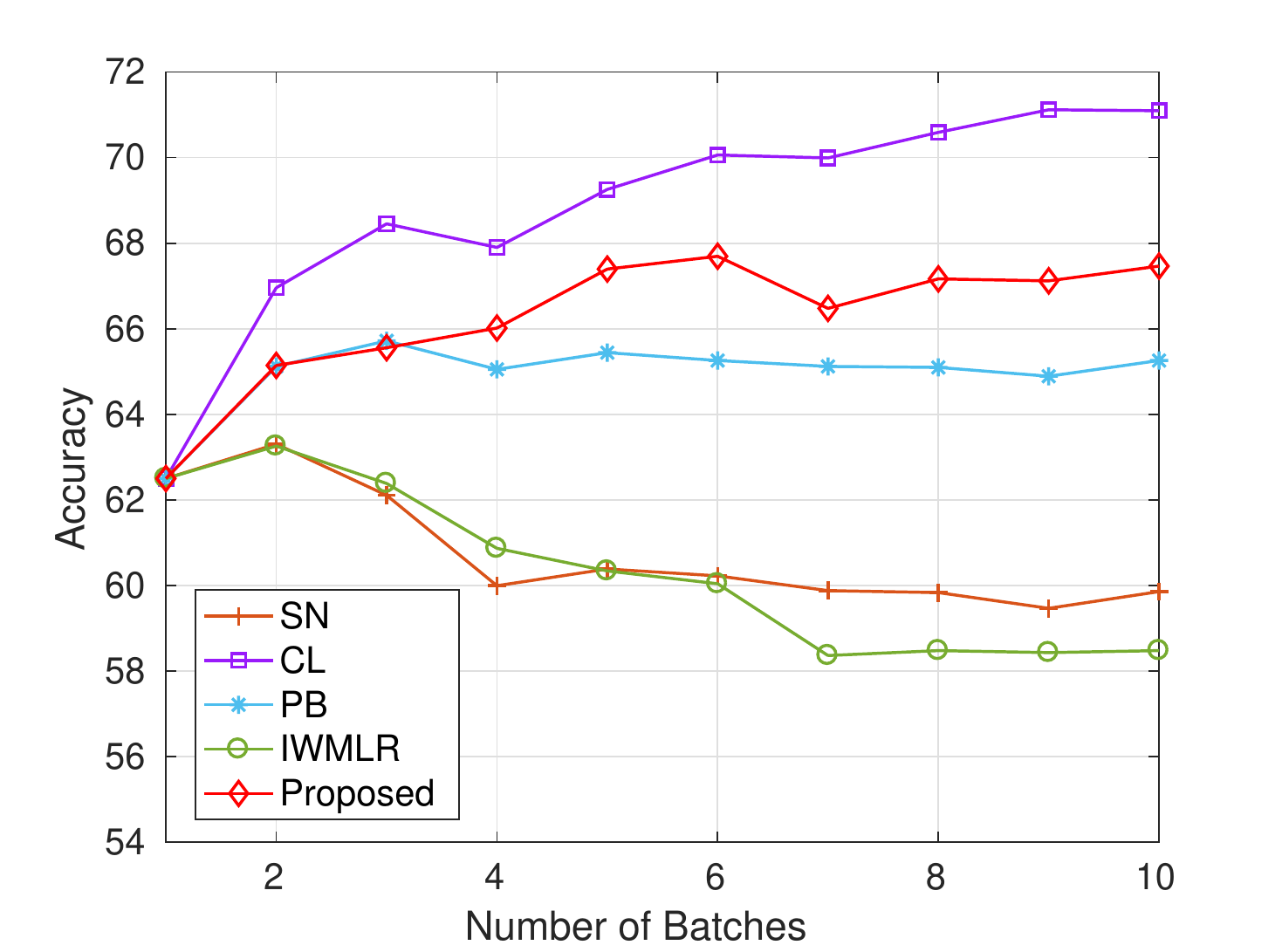}
		\caption{Document}
		\label{acc_40_3}
	\end{subfigure}
	\caption{Comparison of the classification performance of our proposed framework for active learning setup for three different applications: scene classification, activity classification, and document classification. We compare our proposed framework with two baseline approaches (SN, CL) and two noise-robust approaches (PB, IWMLR) for $\Omega=0.40$ and observe superior performance in all three applications.}
	\label{acc_40}
\end{figure*}

\begin{figure*}
	\centering
	\begin{subfigure}{0.3\textwidth}
		\includegraphics[scale=0.38]{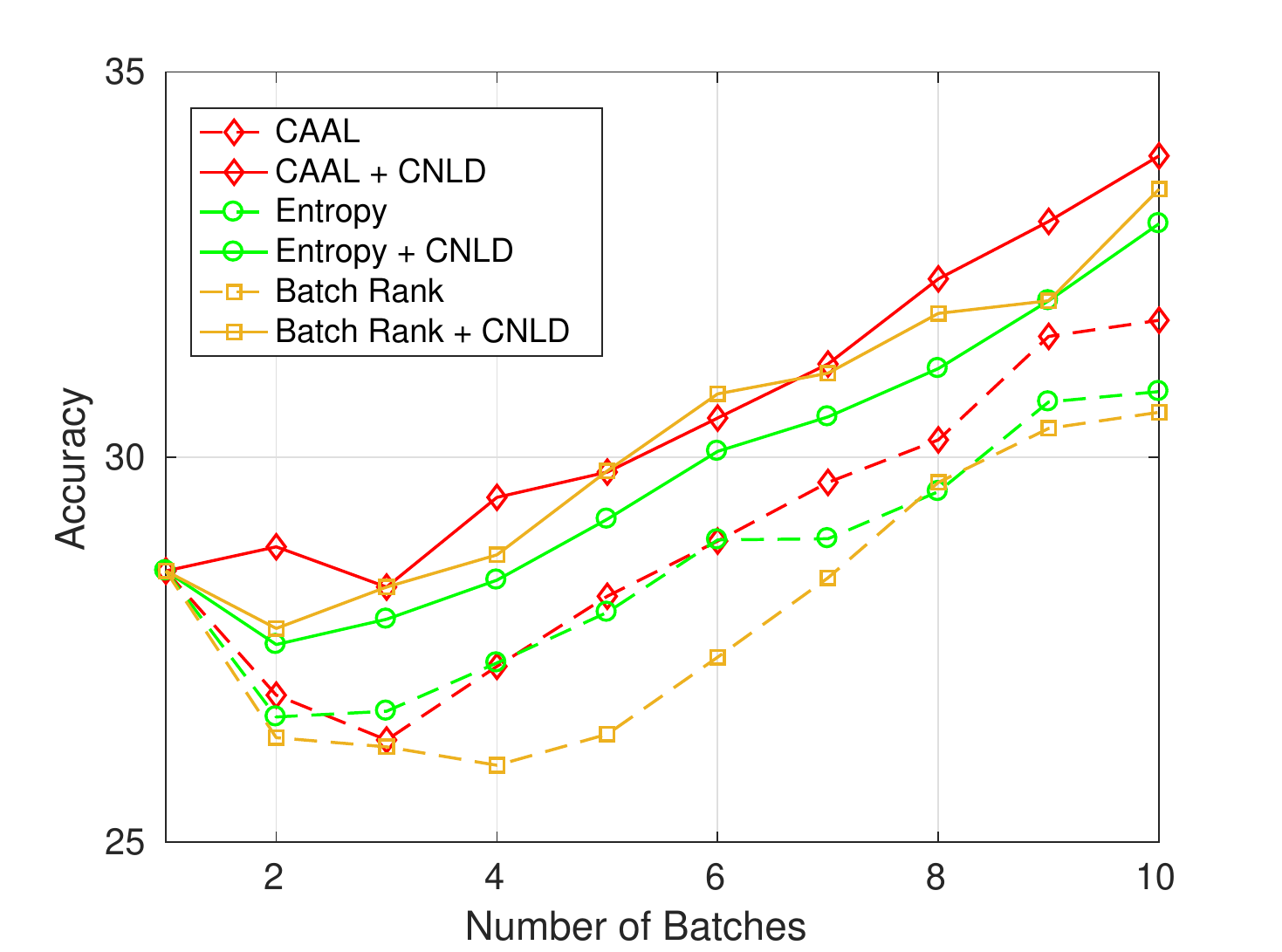}
		\caption{Scene}
		\label{fig:scene_all_20}
	\end{subfigure}
	\begin{subfigure}{0.3\textwidth}
		\includegraphics[scale=0.38]{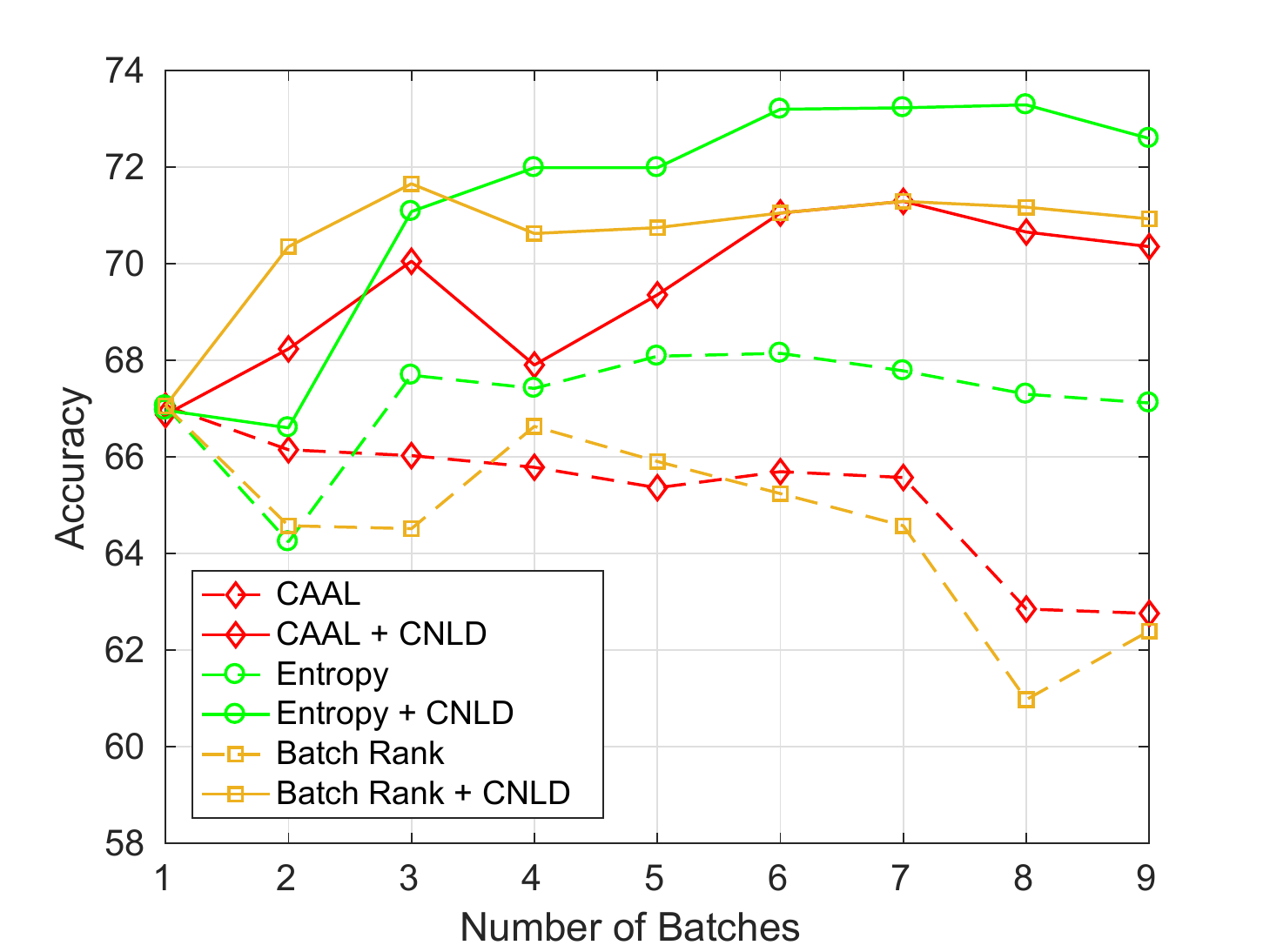}
		\caption{Activity}
		\label{fig:activity_all_20}
	\end{subfigure}
	\begin{subfigure}{0.3\textwidth}
		\includegraphics[scale=0.38]{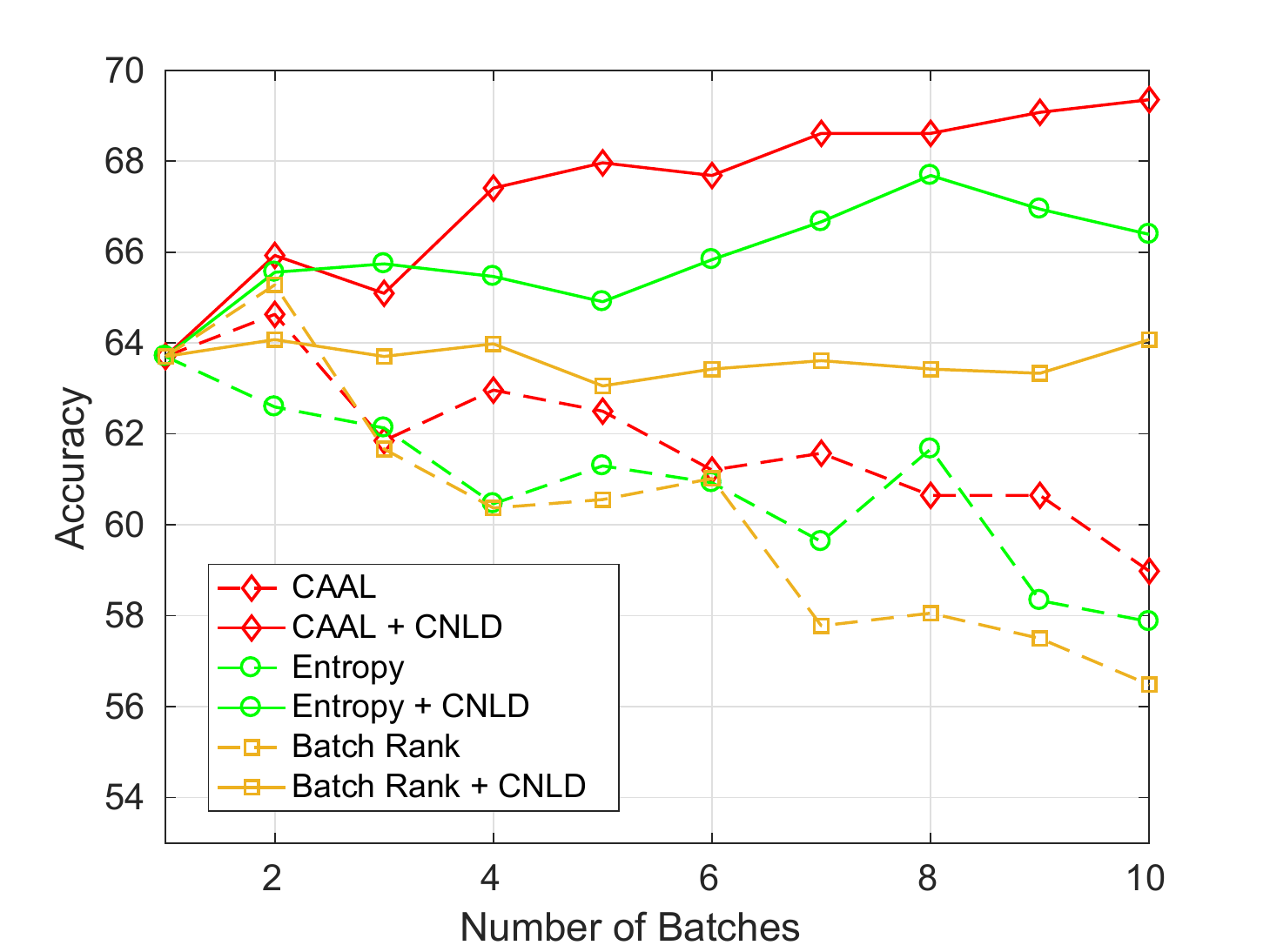}
		\caption{Document}
		\label{fig:doc_all_20}
	\end{subfigure}
	\caption{Analysis of the robustness of three different active learning approaches when combined with the proposed Context-aware Noisy Label Detection (CNLD) framework for $\Omega=0.40$. The figures illustrate that in all three active learning strategies, accuracy is higher when the active learning strategies are combined with CNLD.}
	\label{fig:active_all}
\end{figure*}

\subsection{Classification Robustness Analysis} To analyze the performance of our proposed framework for robust classification in an active learning setup, the entire training set is divided into multiple batches. These batches of data become available sequentially. We assume that all the instances of the initial batch are correctly labeled and data from other batches are unlabeled. The initial batch is used to learn the initial classification model $\mathcal{M}$ and the initial relationship model $\mathcal{R}$. When an unlabelled batch is available, for each unlabeled batch of data, an informative subset of instances is selected and queried for manual labeling. After obtaining the labels from an annotator, a fraction of the annotated instances is randomly selected and assigned wrong labels. To detect the incorrect labels, we represent the data graphically as described in section \ref{FGR} and calculate the dissimilarity scores. Following \ref{sec:MU}, we discard the wrong labels and incrementally update the models with detected correct labels. We evaluate the classification performance on the same testing set whenever the model is updated. We divide the scene and document training set in $10$ batches and the activity training set in $9$ batches. To analyze the performance of our proposed framework, we conduct the following experiments:
\begin{itemize}[leftmargin=*]
    \item We compare the performance of our proposed active learning framework with two baseline approaches and two other label noise-robust approaches for the classification task.
	\item We verify the robustness of different active learning strategies combined with our proposed CNLD approach for the classification task.
	\item We consider a similar incremental setup with pseudo labeling and analyze the impact of CNLD for robust learning.
\end{itemize}

\begin{figure*}
	\centering
	\begin{subfigure}{0.3\textwidth}
		\includegraphics[scale=0.38]{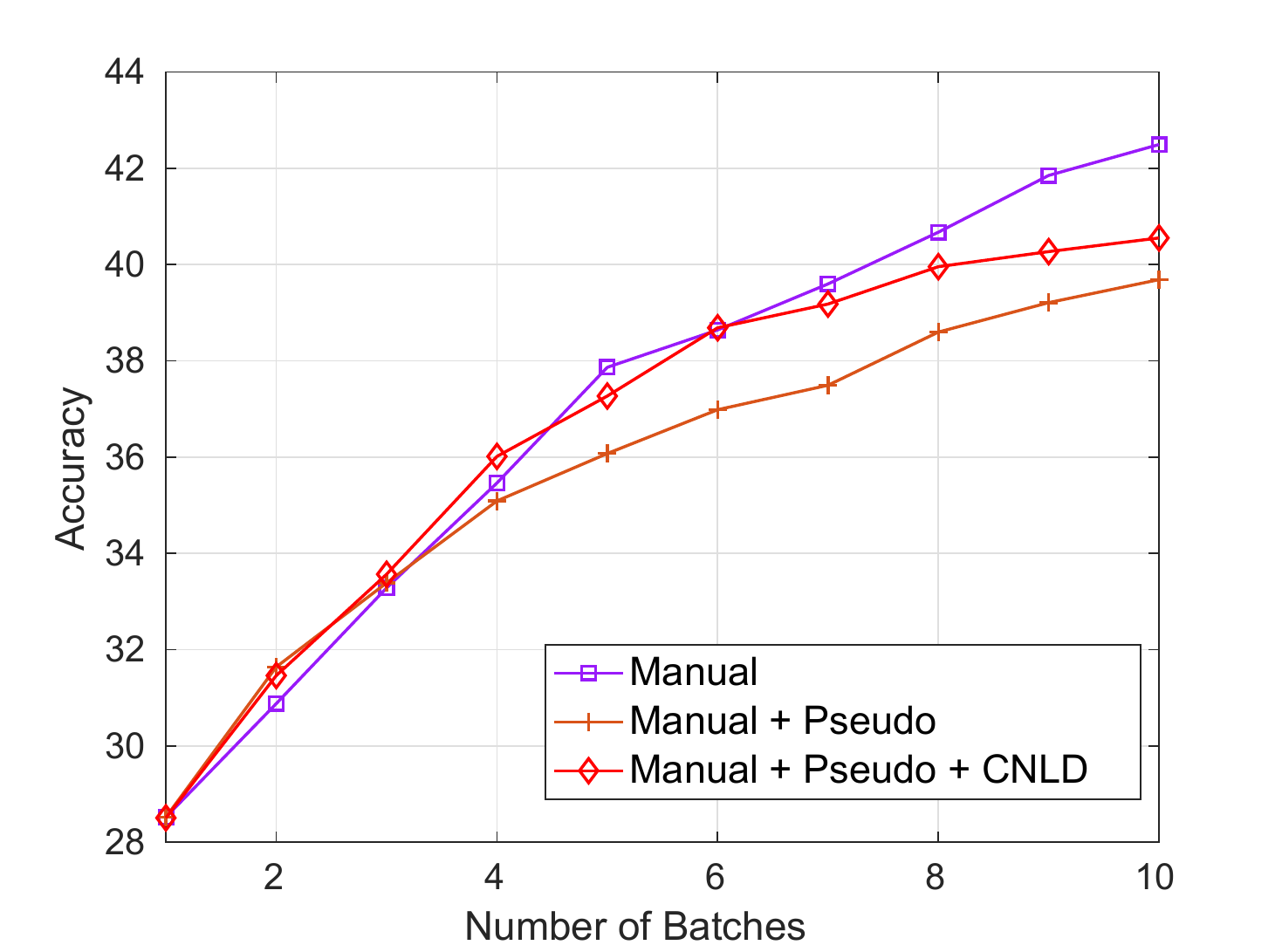}
		\caption{Scene}
		\label{fig:scene_pseudo}
	\end{subfigure}
	\begin{subfigure}{0.3\textwidth}
		\includegraphics[scale=0.38]{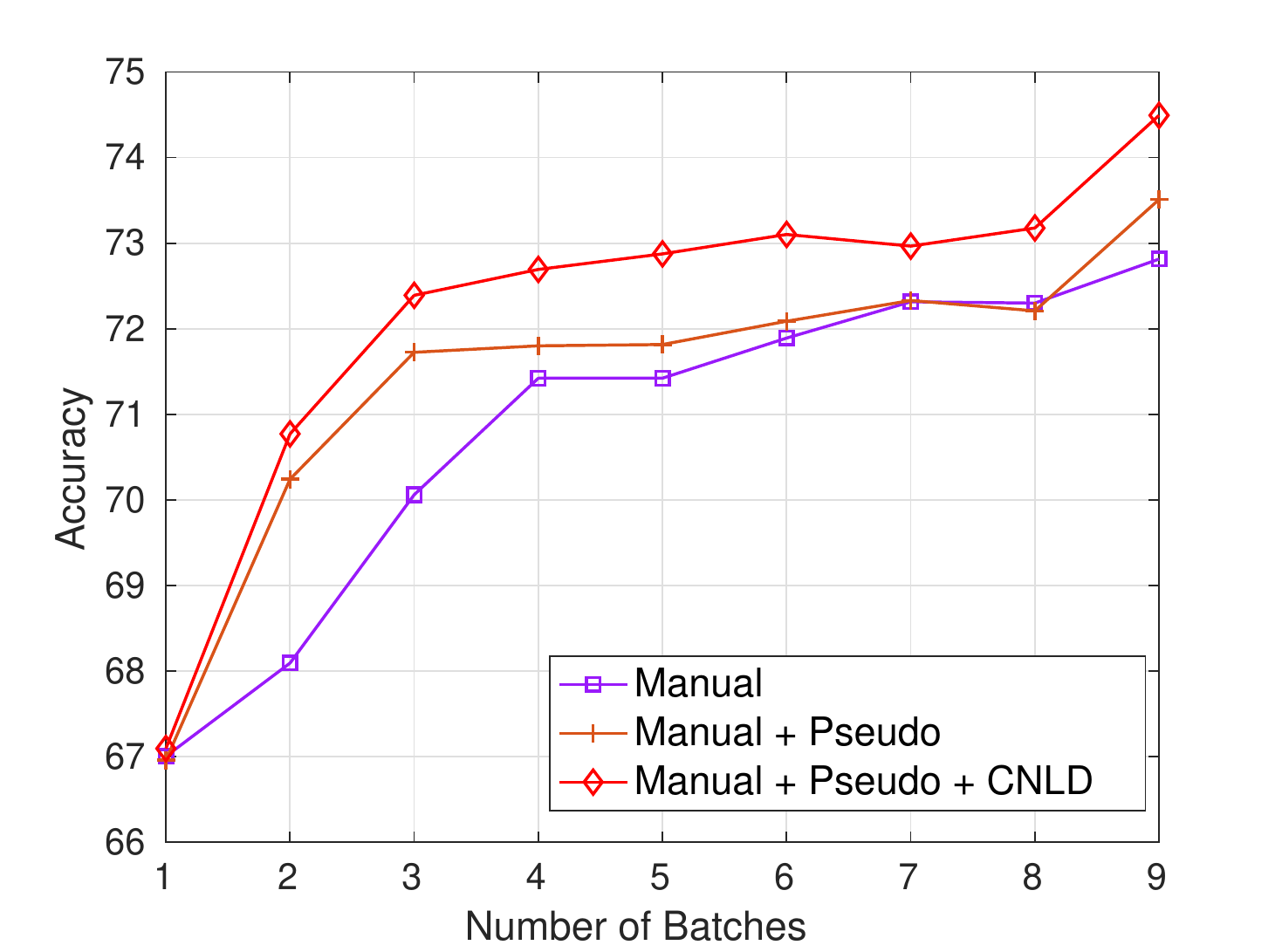}
		\caption{Activity}
		\label{fig:activity_pseudo}
	\end{subfigure}
	\begin{subfigure}{0.3\textwidth}
		\includegraphics[scale=0.38]{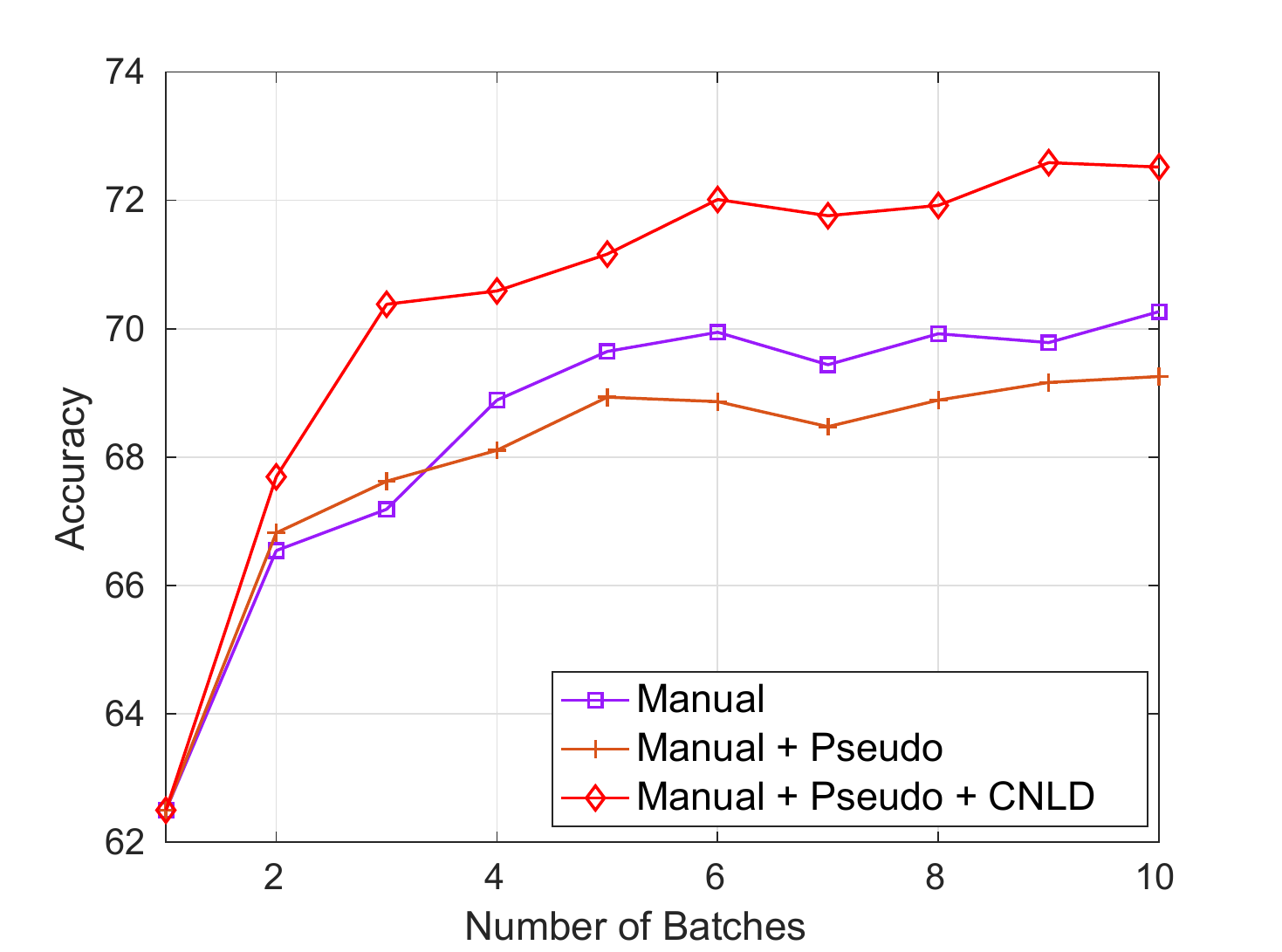}
		\caption{Document}
		\label{fig:doc_pseudo}
	\end{subfigure}
	\caption{Comparison of classification performance with pseudo labels for three different applications: scene, activity, and document classification. We analyse how pseudo labeling combined with CNLD can help improve performance.}
	\label{fig:pseudo}
\end{figure*}

\underline{\textit{Classification Performance Comparison.}} Figure \ref{acc_40} illustrates the classification performance of two baseline approaches and two other label noise-robust approaches compared to our proposed approach. In this experimental setup, we consider symmetric noise with $\Omega = 0.40$ and synthetically assign wrong labels on the queried samples using the NCAR model. We compare our proposed method with the following learning approaches:
\begin{itemize}[leftmargin=*,label=$\diamond$]
    \item \textbf{SN}: In this approach, when the unlabelled batch of data becomes available, the classification model is updated with manually queried labels from the batch. Here, the queried labels are noisy.   
	\item \textbf{PB}: This is also an incremental learning approach. We use the probabilistic approach as discussed in \ref{sec:noise_detection} to detect wrong labels of the queried samples from an unlabelled batch of data and discard them. Then the classification model is updated using the rest of the labels.
	\item \textbf{CL}: In this setup, we utilize the ground truth information of which labels are wrong. The classification model is updated by discarding the wrong labels. This approach represents the upper bound of the classification performance. 
	\item \textbf{IWMLR}: State-of-the-art non-deep learning noise resilient method namely Importance Weighted Multinomial Logistic Regression (IWMLR) \cite{wang2018partial}. Note that IWMLR is proposed for multi-class learning when all the data are available. We adapt it to active learning for proper comparison. When updating a classification model, for each of the queried labels, this method assigns a weight on the sample based on the likelihood of the label is wrong. It enables the learning on noisy data to more closely reflect the results on learning noise-free data.
\end{itemize}

Note that we do not compare with the Multi-class Multi-annotator Robust Gaussian Process (RGP) \cite{long2015multi}. The reasons for this are as follows. i) In the addressed dataset, Gaussian Process Classification (GPC) performs poorly compared to the parametric Multinomial Logistic Regression (MLR). In all three applications, MLR trained with the labels from the initial batch of data performs with higher accuracy compared to GPC. ii) RGP is used in a multi-annotator setting and shown to have a time complexity of $\mathcal{O}(n^3)$ \cite{hernandez2011robust}, which makes the approach not suitable for the addressed datasets. In PB and CL approach, we discard the same number of labels as our proposed approach to make a fair comparison.

For all three classification tasks, SN illustrates the negative consequence of label noise on the learning process. CL illustrates the upper bound of the classification performance by discarding the wrong labels. We do not observe much improvement for IWMLR compared to SN for these datasets. Compared to the scene classification task, PB performs better than SN in both activity and document classification tasks. This is apparent as the noisy label detection performance of the probabilistic approach is better for activity and document classification. However, the plots in Fig. \ref{acc_40} illustrate that during the active learning process, our proposed method retains better accuracy than other approaches for each stage of updating the model with a new set of noisy labels. We provide the performance comparison of our proposed approach with SN for $\Omega \in \{0.20,0.30,0.50\}$ in the supplementary material.

\underline{\textit{Robustness of Different Active Learning Strategies.}} To demonstrate that the proposed framework is independent of the selection of an active learning strategy, we analyze the robustness of different active learning techniques when combined with CNLD. Figure \ref{fig:active_all} illustrates the robustness of our proposed approach for different active learning strategies. We select three commonly used active learning methods: Entropy \cite{li2013adaptive}, Batch Rank \cite{chakraborty2015active}, and CAAL \cite{hasan2018context}. For each of the active learning approaches, we compare the performance of the classifier when updated with noisy labels vs. when updated with detected correct labels by CNLD. In Figure \ref{fig:active_all}, the dotted lines refer to the classification performance of different active learning strategies where the classifier is updated with noisy labels. The solid lines in Figure \ref{fig:active_all} represent the classification performance when the active learning approach is combined with CNLD. For all three applications and for each of the active learning approaches, frameworks combined with CNLD are more robust and result in higher accuracy compared with their vanilla counterpart. 

\begin{table}[t]
\centering
\caption{In this table, we report the improvement in performance for our proposed approach over SN for different selection of $\beta$. We show the results for $\Omega\in\{0.10,0.20,0.30,0.40,0.50\}$ and for different $\beta$ selection from the set $\{ 0.80,0.85,0.90\}$.}
\label{table:beta}
\begin{tabular}{c|c|ccc}
\hline
& \multirow{2}{*}{Noise} & \multicolumn{3}{c}{\underline{Accuracy improvement over $SN$}} \\
& & $\beta=0.80$ & $\beta=0.85$ & $\beta=0.90$\\
\hline
\hline
\parbox[c]{2mm}{\multirow{5}{*}{\rotatebox[origin=c]{90}{Scene}}} & $\Omega=0.10$ & $0.24\pm0.47$ & $0.03\pm0.44$ & $-0.29\pm0.83$ \\
& $\Omega=0.20$ & $0.36\pm0.39$ & $0.22\pm0.71$ & $0.06\pm0.71$\\
& $\Omega=0.30$ & $1.02\pm0.70$ & $1.00\pm0.61$ & $1.51\pm0.80$\\ 
& $\Omega=0.40$ & $1.93\pm0.87$ & $1.78\pm0.90$ & $1.65\pm0.70$\\ 
& $\Omega=0.50$ & $1.99\pm0.90$ & $2.46\pm1.13$ & $3.81\pm1.70$\\
\hline
\hline
\parbox[c]{2mm}{\multirow{5}{*}{\rotatebox[origin=c]{90}{Activity}}} & $\Omega=0.10$ & $0.54\pm0.53$ & $0.62\pm0.59$ & $1.23\pm0.72$\\  
& $\Omega=0.20$ & $-0.05\pm0.79$ & $0.34\pm0.60$ & $0.26\pm0.59$\\
& $\Omega=0.30$ & $1.21\pm0.84$ & $1.44\pm1.03$ & $2.43\pm1.76$\\ 
& $\Omega=0.40$ & $1.68\pm1.57$ & $1.57\pm1.02$ & $3.45\pm2.21$\\ 
& $\Omega=0.50$ & $1.50\pm1.54$ & $2.40\pm1.60$ & $4.25\pm3.03$\\
\hline
\hline
\parbox[c]{2mm}{\multirow{5}{*}{\rotatebox[origin=c]{90}{Document}}} & $\Omega=0.10$ & $1.09\pm0.94$ & $0.67\pm0.82$ & $1.43\pm1.25$\\  
& $\Omega=0.20$ & $1.13\pm0.72$ & $1.57\pm0.85$ & $1.76\pm0.98$\\
& $\Omega=0.30$ & $3.18\pm1.78$ & $2.96\pm1.64$ & $4.06\pm2.14$\\ 
& $\Omega=0.40$ & $3.04\pm1.49$ & $5.06\pm2.71$ & $5.55\pm2.79$\\ 
& $\Omega=0.50$ & $3.72\pm1.84$ & $4.82\pm2.32$ & $6.51\pm3.44$\\
\hline
\end{tabular}
\end{table}

\underline{\textit{Pseudo Labeling}} We consider another incremental setup where we update the models with manually queried labels and pseudo labels. In this setup, the initial models are learned using the initial set of data. When a new batch of data becomes available, we select an informative set of samples and query the labels. We consider these queried labels to be correct. We also utilize the unlabelled data and update the classification models with predicted labels of the unlabelled data, which is called the pseudo labels. The generation of pseudo labels is classifier dependent and can contain a lot of noise. Here, the generation of noise is feature dependent and more closely reflects the real noise scenario. So while updating models with pseudo labels, we use CNLD to detect the wrong labels and filter them. In this experimental setup, we use three learning approaches and compare their performance:
\begin{itemize}[leftmargin=*]
    \item \textbf{Manual}: In this approach, when an unlabelled batch of data is available, we update the models with only manually queried correct labels.  
    \item \textbf{Manual + Pseudo}: In this approach, when an unlabelled batch of data is available, we update the models with manually queried correct labels and generated pseudo labels from the rest of the unlabelled data of that batch. Here, the pseudo labels are generated using the current classification model.  
    \item \textbf{Manual + Pseudo + CNLD}: Similar to the setup of Manual + Pseudo. Additionally, we consider utilizing CNLD to identify pseudo labels that are wrong. Then the model is updated with manually queried correct labels and pseudo labels that are detected as correct. 
\end{itemize}

Figure \ref{fig:pseudo} illustrates the performance comparison of the above-mentioned learning approaches. In the scene classification task, the model updated with manually correct labels (Manual) performs best. This is because the accuracy of the learned classifier is low and generates a lot of wrong labels, which eventually degrade the classification performance if used for updating the model.  However, compared to using pseudo labels directly (Manual + Pseudo), there is an improvement in performance if we filter wrong pseudo labels using CNLD (Manual + Pseudo + CNLD). In activity and document classification task, Manual performs better than Manual + Pseudo. However, in both classification tasks, performance of the filtered pseudo labeling approach (Manual + Pseudo + CNLD) is superior to Manual and Manual + Pseudo.

\underline{\textit{$\beta$ Parameter Sensitivity Analysis.}} The selection of $\beta$ parameter is a trade-off between precision and recall for noisy label detection. A process with high precision may not be able to detect a lot of incorrect labels. On the other hand, a process with high recall will detect a lot of correct labels as wrong. In both cases, the performance of a classification model will degrade. As a result, our approach requires to select $\beta$ in a way to balance between these two conditions. In Table \ref{table:beta}, we analyze the performance of the classification model for a wide range of selection of $\beta$. For noise rate $\Omega\in\{0.10,0.20,0.30,0.40,0.50\}$ and $\beta\in\{0.80,0.85,0.90\}$, we report the absolute classification accuracy improvement of our proposed approach over the baseline SN approach. We compute the average of the difference of accuracy of our proposed approach and SN approach in each incremental update and also report the standard deviations. We observe that for a wide range of selection of $\beta$, our approach gains a performance improvement over learning with noisy labels. Improvement of performance is not significant for a low noise rate ($\Omega\in\{0.10, 0.20\}$). It is expected because a small number of noisy labels do not degrade the classification performance much. For $\Omega \in \{0.30, 0.40, 0.50\}$, there is $1\% - 3.81\%$ absolute accuracy improvement in scene classification, $1.21\% - 4.25\%$ absolute accuracy improvement in activity classification, and $3.18\% - 6.51\%$ absolute accuracy improvement in document classification for different $\beta$ selection.

\begin{figure}
    \begin{subfigure}{0.3\textwidth}
	\includegraphics[scale=0.313]{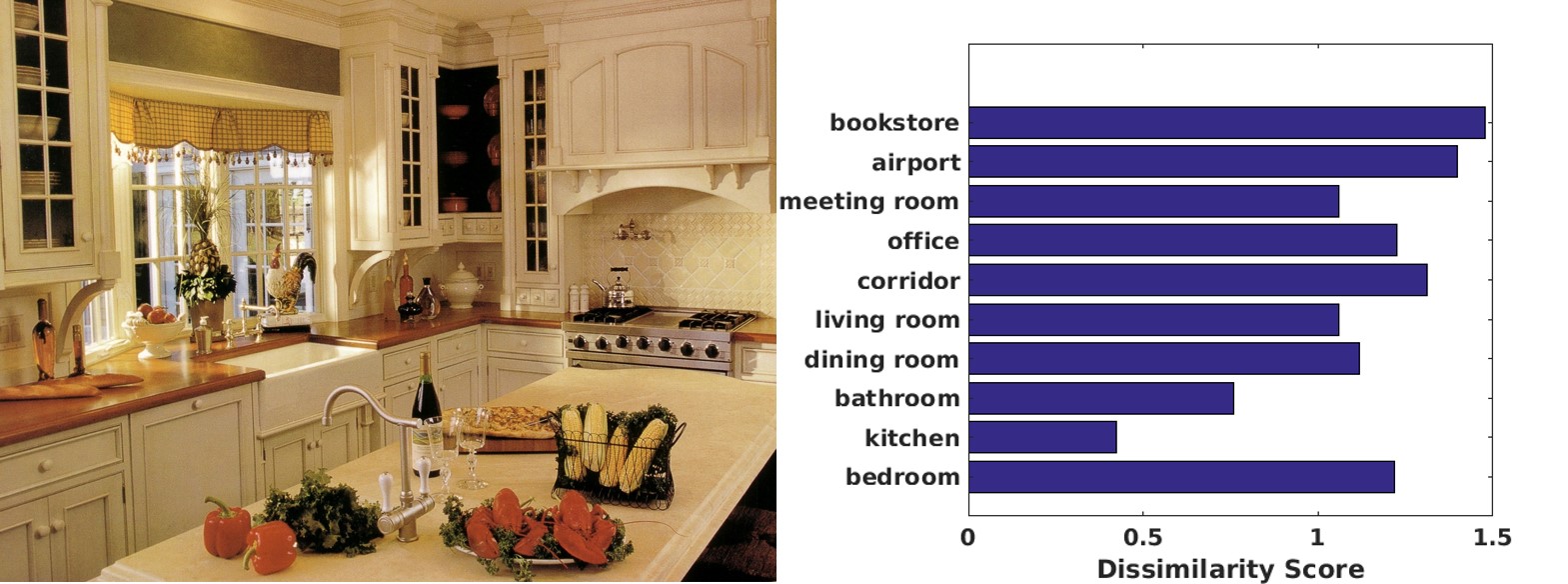}
	\label{fig:example_a}
	\end{subfigure}
	\begin{subfigure}{0.3\textwidth}
	\includegraphics[scale=0.31]{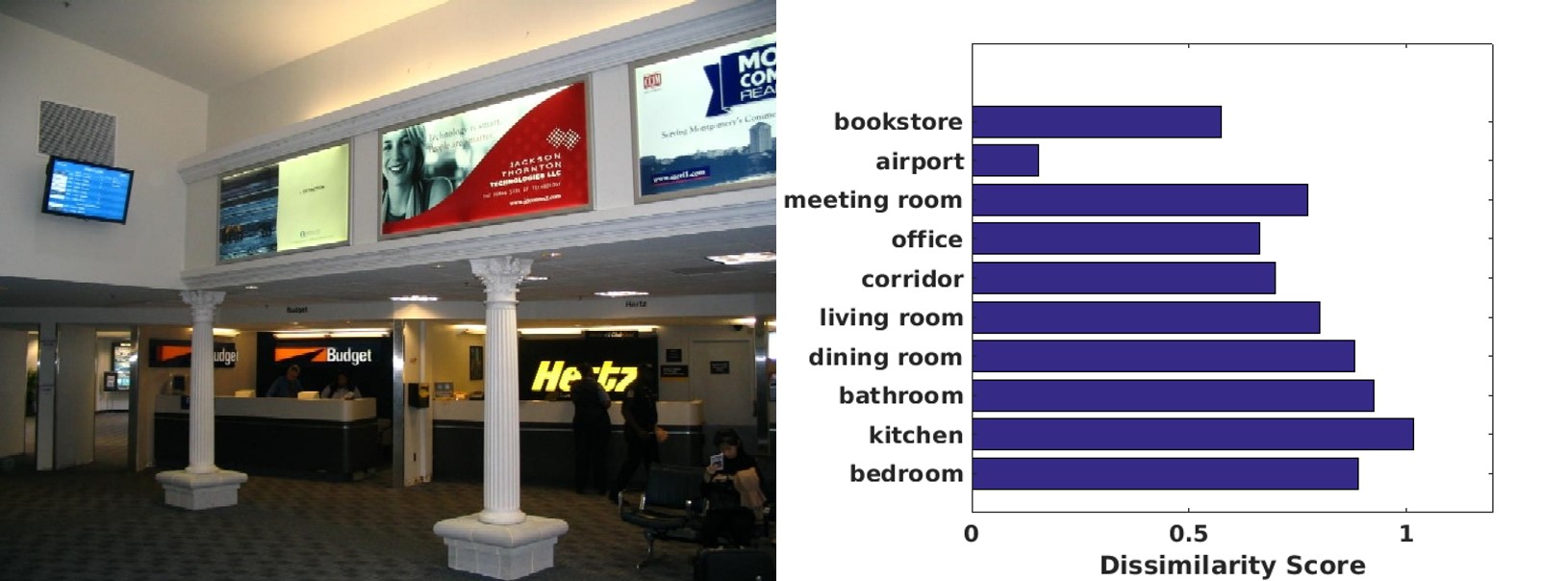}
	\label{fig:example_b}
	\end{subfigure}
	\caption{Example illustration of the performance of noisy label detection. The dissimilarity score is minimum when the first image is labeled with the correct scene class `kitchen'. Similarly, the dissimilarity score for the second image is minimum when the image is assigned with the `airport' scene class.}
	\label{fig:qualitative}
\end{figure}

\subsection{Qualitative Results} We provide two qualitative examples from the MIT-67 Indoor dataset for the noisy label detection task in Figure \ref{fig:qualitative}. The graphical representation is constructed using the objects that are detected by an off-the-shelf object detector. Then we utilize the graphical representation and the prior relational information to compute the dissimilarity scores. A high dissimilarity score indicates the label is likely to be wrong, while a low dissimilarity score indicates the label is correct. For example, when the top image in Figure \ref{fig:qualitative} is labeled as `kitchen' scene, the contextual relation formed with object labels and a scene label will be consistent with previously learned contextual relation. As a result, the dissimilarity score is low for `kitchen' class. If the image is assigned with any other label except for `kitchen', there will be a contextual inconsistency and the dissimilarity score will be high. Similarly, for `airport' indoor image, for `airport' label the dissimilarity score is low while for other assigned labels, the dissimilarity score is high. 

%% file: sections/N06_conclusion.tex
\section{Conclusion}
In this paper, we formalize a general active learning framework that utilizes a noisy label filtering based learning approach to reduce the adverse impact of label noise. In this regard, we propose a novel context-aware noisy label detection strategy. For various applications, we show how we can represent the inter-relationship among the data and using that representation, infer the likelihood of a label being noisy. The proposed noisy label robust framework is independent of a particular choice of feature, classifier, and active selection strategy. We experimentally validate the robustness of the active learning approach in the presence of label noise.